\documentclass[lettersize,journal]{IEEEtran}
\usepackage{amsmath,amsfonts}
\usepackage{algorithmic}
\usepackage{algorithm}
\usepackage{array}
\usepackage[caption=false,font=normalsize,labelfont=sf,textfont=sf]{subfig}
\usepackage{textcomp}
\usepackage{stfloats}
\usepackage{url}
\usepackage{verbatim}
\usepackage{graphicx}
\usepackage{cite}
\usepackage{bm}
\usepackage{multirow}

\begin{document}

\title{Boundary Voting Network for Ambiguity-aware Timestamp-supervised Action Segmentation}

\author{
    Runzhong Zhang,
    \and
    Yueqi Duan,~\IEEEmembership{Member,~IEEE,}
    \and
    Yang Chen,
    \and
    Weipeng Hu, \\
    \and
    Chen Cai,
    \and
    Suchen Wang,
    \and
    and Yap-Peng Tan,~\IEEEmembership{Fellow,~IEEE} \\
\thanks{Runzhong Zhang, Yang Chen, Weipeng Hu, Chen Cai, and Yap-Peng Tan are with the School of Electrical and Electronic Engineering, Nanyang Technological University, Singapore (e-mail: runzhong001@e.ntu.edu.sg, chen1605@e.ntu.edu.sg, weipeng.hu@ntu.edu.sg, e190210@e.ntu.edu.sg, eyptan@ntu.edu.sg).}
\thanks{Yueqi Duan is with the Department of Electronic Engineering, Tsinghua University, Beijing, China (e-mail: duanyueqi@tsinghua.edu.cn).}
\thanks{Suchen Wang is with the Amazon, Seattle, America (e-mail: wansuche@amazon.com).}
\thanks{Corresponding author: Yueqi Duan.}
}

\markboth{IEEE TRANSACTIONS ON CIRCUITS AND SYSTEMS FOR VIDEO TECHNOLOGY, VOL. 35, NO. 11, NOVEMBER 2025}%
{Shell \MakeLowercase{\textit{et al.}}: A Sample Article Using IEEEtran.cls for IEEE Journals}

\IEEEpubid{1051-8215~\copyright~2025 IEEE.}

\maketitle

\begin{abstract}
Timestamp-supervised action segmentation aims to segment and classify actions in untrimmed videos with a random frame annotated per action. Precisely localizing action boundaries from timestamp annotations is crucial for this setting, as it enables generating framewise pseudo-labels and applying the well-explored fully-supervised training. However, prevailing methods struggle with intrinsic uncertainty in boundary localization due to less discriminative features in action-transiting regions. This imprecise boundary estimation significantly reduces the stability and reliability of the generated pseudo-labels in ambiguous action-transiting regions, consequently resulting in performance deterioration of the trained segmentation models. In our paper, we introduce the boundary voting network that mitigates feature ambiguity by hierarchically propagating video-level global prior knowledge into local action-transiting regions. By generating key action representations as votes throughout the video and targeting action-transiting regions, all votes collaboratively contribute to action-transiting feature enhancement and boundary localization refinement. Extensive experiments demonstrate the effectiveness of our method on GTEA, 50Salads, and Breakfast datasets.
\end{abstract}


\section{Introduction}
\IEEEPARstart{A}CTION segmentation aims to temporally segment untrimmed video sequences by assigning each frame a pre-defined action class~\cite{li2023involving}. The methodology has emerged as a promising approach for understanding long-form videos with complex action content~\cite{li2024neighbor, cheng2024continual, cheng2025achieving}, holding profound significance for applications such as human-robot interactions~\cite{cui2023active, cai2024empowering, zhang2023hoi, yang2021rr, wang2023exploring}, video captioning~\cite{yan2022video, wang2020event, qi2019sports, cai2024temporal, xu2022bridging}, and advanced home monitoring systems~\cite{dai2022toyota, wang2020smrt, bao2023cross, bao2024omnipotent}. This paper focuses on the timestamp-supervised action segmentation setting, which involves annotating only a single, random frame for each action segment within the training video.

The key to the setting is inferring precise boundary localization from timestamp annotations, as it enables the generation of framewise pseudo-labels, which effectively transforms timestamp supervision into the well-explored fully-supervised problem~\cite{behrmann2022unified, khan2022timestamp, li2021temporal, zhao2022turning}. To elaborate, by localizing the action boundary between every two consecutive annotated timestamps, the preceding annotation class can be automatically assigned to every frame before the boundary, while the subsequent annotation class to frames after the boundary. This iterative procedure, applied to each annotated timestamp pair, generates framewise pseudo-labels spanning the entire training video and facilitates fully-supervised learning. Consequently, the effectiveness of boundary localization directly influences the gap between timestamp and upper-bound full supervision, playing a pivotal role in both the model learning process and segmentation testing performance.

\begin{figure*}[t]
 \centering
 \includegraphics[width=0.92\linewidth]{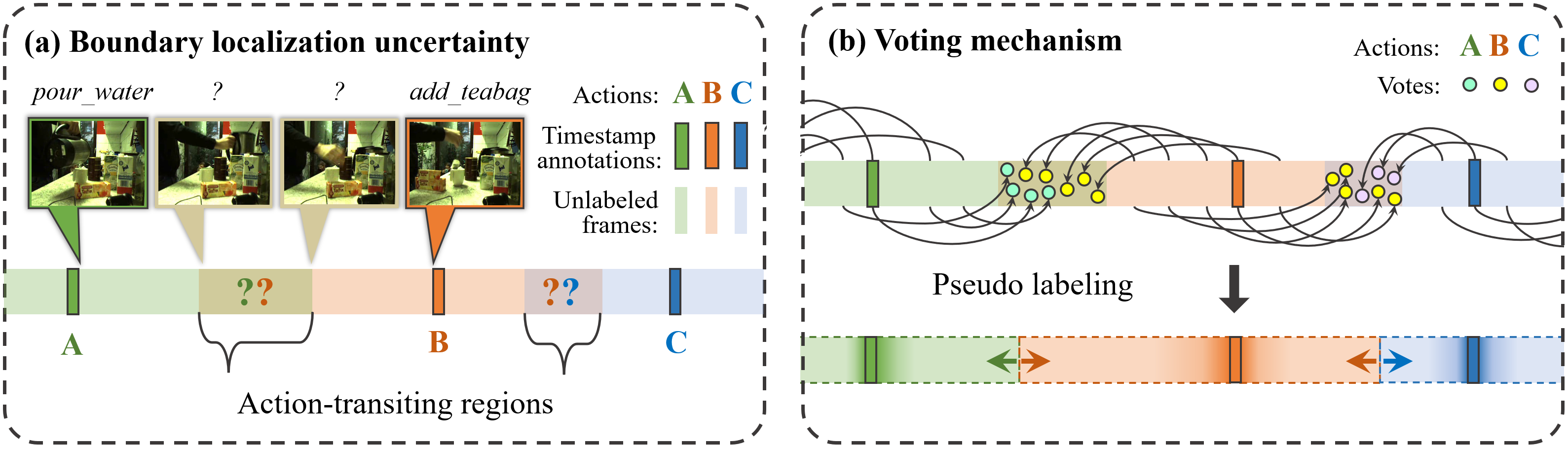}
  \caption{(a) Due to the inherent feature ambiguity in action-transiting regions, existing works suffer from significant uncertainty in boundary localization, resulting in inaccurate framewise pseudo-label generation and cascaded performance deterioration. (b) Our network propagates and aggregates key action representations as votes into the action-transiting regions, thereby refining boundary localization and improving segmentation performance in both the training and testing phases.}
  \label{fig:teaser}
\end{figure*}

Despite the notable progress achieved, most prior works encounter heightened uncertainty in boundary localization due to feature ambiguity in action-transiting regions. The transitional frames, which occur adjacent to the boundaries where actions shift, often exhibit less discriminative semantic representations compared to frames indicating the progression of actions. For example, in the tea-making video shown in Fig.~\ref{fig:teaser} (a), the transition from ``pouring water" to ``adding teabag" involves placing down the pot and moving the hand from the pot to the teabag, rather than explicit interactions with the objects. Moreover, the long duration of actions in the video often places boundaries at considerable temporal distances from any action centroid, further diminishing the correlation between transitional frames and other parts of the video containing rich action information. Consequently, the feature ambiguity in action-transiting regions leads to a lack of essential knowledge for precise boundary localization and instability in the generated pseudo-labels, resulting in cascaded performance deterioration in the subsequent fully-supervised training phase.

\IEEEpubidadjcol
In contrast to previous studies primarily focusing on model-level architecture adjustments~\cite{behrmann2022unified, li2021temporal, zhao2022turning}, we reconsider boundary localization uncertainty through the feature-level enhancement strategy. Our proposed boundary voting network (BVN), depicted in Fig.~\ref{fig:teaser} (b), introduces a global-to-local voting mechanism that enhances local action-transiting features by integrating global prior knowledge across the video. Specifically, we first generate intermediary framewise features using the segmentation model encoder and identify action-transiting regions through bidirectional boundary detection, following the procedure in~\cite{li2021temporal}. Due to the feature ambiguity surrounding boundaries, action-transiting regions persist for long temporal duration, leading to inaccurate framewise pseudo-label generation. To address this, we introduce two modules named the voting block and the aggregation block. The voting block derives two frame features as votes from each intermediary framewise feature, targeting the center of the corresponding action-transiting regions, with one toward the start of the action and the other toward its end. Subsequently, the aggregation block merges the votes back into the original video features, thereby enhancing action representations within action-transiting regions. Consequently, the proposed blocks globally generate and rearrange key action representations from the evenly distributed temporal sequence, with the new characteristic of clustering around boundaries, which facilitates feature enhancement in targeted regions. By hierarchically applying blocks among segmentation decoding layers, BVN continuously mitigates feature ambiguity and suppresses the action-transiting regions during training, ultimately generating more accurate framewise pseudo-labels and improving segmentation performance in both training and testing phases.

To the best of our knowledge, our main contributions are summarized as follows:

\begin{itemize}
\item  We propose the first work to investigate the inherent feature-level ambiguity in action-transiting regions and address boundary localization uncertainty through a voting mechanism.

\item The global-to-local BVN sequentially constructs the voting block to propagate global prior knowledge towards local action-transiting regions, and the aggregation block to enhance targeted less discriminative features. Our model hierarchically suppresses ambiguous regions and accurately separates adjacent actions during both training and testing phases.

\item Our method demonstrates its superiority across three real-world datasets with various evaluation metrics. Additionally, we validate the effectiveness of the voting mechanism through comprehensive ablation studies. 
\end{itemize}

\section{Related Works}
In this section, we briefly review recent works on fully-supervised action segmentation and timestamp-supervised action segmentation.

\subsection{Fully-supervised action segmentation}
Existing fully-supervised approaches segment videos with dense framewise annotations provided during training. Early approaches~\cite{karaman2014fast,rohrbach2012database} classified frames through sliding windows with non-maximum suppression, which failed to capture the temporal dependency between actions. To model action sequences, alternative approaches applied Markov model~\cite{lea2016segmental} and recurrent neural network~\cite{yeung2018every}. Recently, various temporal convolutional network-based approaches have been proposed to capture long-range action dependencies effectively. Lea \textit{et al}.~\cite{lea2017temporal} introduced the use of TCN in action segmentation by designing an encoder-decoder architecture with 1D temporal convolution/deconvolution. Ding \textit{et al}.~\cite{ding2017tricornet} presented a hybrid network that integrates RNN into TCN for the decoder layers, and TDRN~\cite{lei2018temporal} replaced temporal convolution with deformable convolution. However, these methods all down-sampled the video and inevitably led to information loss. Instead, Farruha \textit{et al}.~\cite{farha2019ms} proposed MS-TCN, a multi-stage temporal convolutional network processing the video at full length. In MS-TCN++, Li \textit{et al}.~\cite{li2020ms} introduced dual dilated temporal convolution to capture local-global features and improved the model efficiency. Beyond TCN frameworks, recent efforts have explored diverse model architectures and training strategies. ASFormer~\cite{yi2021asformer} introduced the first transformer-based network for action segmentation. Li \textit{et al}.~\cite{li2022bridge} utilized the textual information in action labels and contrastively trained the text encoder with the video encoder. DiffAct~\cite{liu2023diffusion} integrated diffusion into action segmentation for the first time. FACT~\cite{lu2024fact} introduced a two-branch architecture that concurrently learns both frame-level and action-level features, and the communication mechanism between the branches through cross-attention. However, while fully-supervised approaches have demonstrated remarkable performance across various datasets, obtaining framewise labels demands significant manual effort, particularly for large-scale real-world video data.

\begin{figure*}[t]
 \centering
 \includegraphics[width=1.0\linewidth]{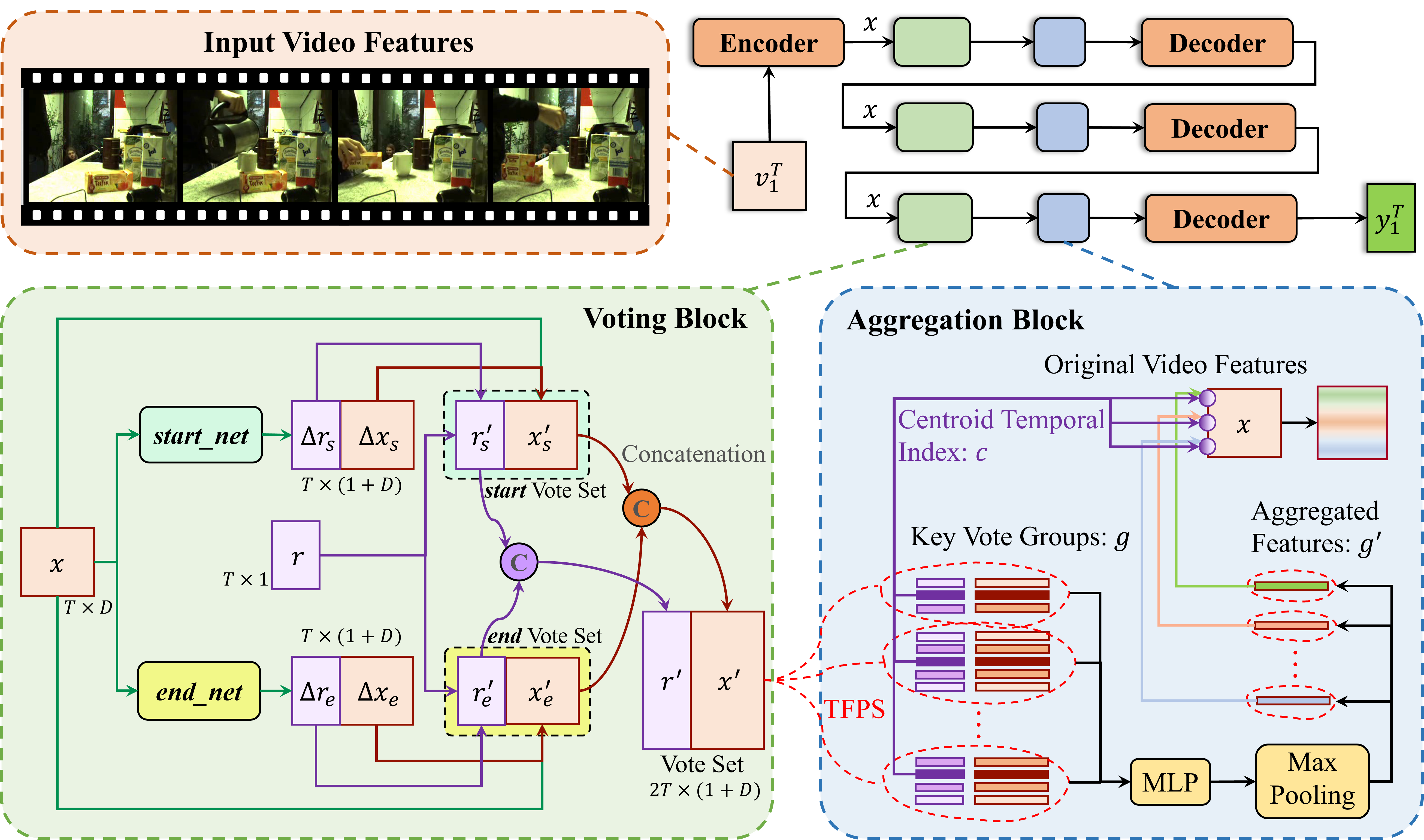}
  \caption{Overview of the global-to-local BVN architecture. After obtaining video features $x$ from the encoder, our model iteratively processes them through stages containing the voting block, aggregation block, and decoder. Taking $x$ and normalized temporal indexes $r$ as inputs, the voting block computes features $x_s'$ and $x_e'$ to capture key action representations, and predicts the start time $r_s'$ and end time $r_e'$ of the current action independently through $start\_net$ and $end\_net$. Based on the obtained vote set $\left\{x', r' \right\}$, the subsequent aggregation block constructs temporal farthest point sampling (TFPS) to select representative votes, which are then aggregated into the original video features for enhancement. On top of the final decoder outputs, we calculate the framewise predictions $y^T_1$ using MLP. Through hierarchical voting propagation and aggregation across stages, our model continuously suppresses action-transiting feature ambiguity while simultaneously refining boundary localization.}
  \label{fig:network}
\end{figure*}

\subsection{Timestamp-supervised action segmentation}
For the timestamp-supervised action segmentation, each action segment within the training video is annotated using a single arbitrary frame. Inspired by the foundational concept of point-supervised semantic segmentation~\cite{bearman2016s}, Li \textit{et al}.~\cite{li2021temporal} introduced a learning strategy for timestamp annotations that first generate framewise pseudo-labels through action boundary estimation and proceed with fully-supervised training. Advancing beyond the approach, Khan \textit{et al}.~\cite{khan2022timestamp} replaced heuristic boundary estimation with the graph neural network~\cite{kipf2016semi}, enabling learning in an end-to-end manner. To reduce oscillations during training, Zhao \textit{et al}.~\cite{zhao2022turning} proposed a teacher network in parallel with the segmentation model. In UVAST~\cite{behrmann2022unified}, constrained K-medoid was directly applied to input video features, offering an alternative strategy distinct from inferring boundaries on intermediary features. Instead of assigning framewise hard labels, Rahaman \textit{et al}.~\cite{rahaman2022generalized} acknowledged label uncertainty in unlabeled frames and employed an Expectation-Maximization approach, demonstrating remarkable robustness in handling annotation errors. Recently,  Liu \textit{et al}.~\cite{liu2023reducing} proposed D-TSTAS to reduce over-reliance on annotated timestamps, which ensured the model captured more contextual information in both initialization and refinement steps. Nevertheless, despite extensive efforts directed toward the timestamp-supervised setting, the boundary localization uncertainty associated with feature ambiguity has yet to be thoroughly explored.

\section{Boundary voting network}
The objective of action segmentation is to segment the video into distinct procedural steps and classify human action within each segment. This study focuses on the timestamp-supervised setting, which relies on annotations for only a single random frame per action instead of framewise labels. For a training sample $v^T_1=[v_1, ..., v_T]$ of length $T$ containing $N$ action segments, we employ sparse annotations in the form of $\left\{l^N_1, a^N_1\right\}$, where $l^N_1=[l_1, ..., l_N]$ denotes the timestamps of annotated frames, and $a^N_1=[a_1, ..., a_N]$ indicates the corresponding action labels, with $N$ being significantly lower than $T$. Our goal is to train a model using sparsely annotated timestamps capable of accurately classifying each video frame, represented as $y^T_1=[y_1, ..., y_T]$. It is important to note that the values of $T$ and $N$ may vary for each video.

\subsection{Overview}
In the timestamp-supervised setting, the primary challenge lies in boundary localization uncertainty caused by feature ambiguity, as shown in Fig.~\ref{fig:teaser}. Inspired by conventional Hough voting~\cite{leibe2008robust} in object detection, we introduce a global-to-local boundary voting network to propagate and aggregate key action representations into action-transiting regions, thereby enhancing the less discriminative features within local regions from a global perspective. As illustrated in Fig.~\ref{fig:network}, we construct an encoder-decoder action segmentation network based on the transformer architecture~\cite{vaswani2017attention}. The encoder processes video features $v^T_1$ to generate intermediary framewise features $x$. To effectively propagate action representations in the temporal domain, we introduce a voting block along with a votes aggregation block seamlessly integrated before each decoder layer. Given that the inputs of every voting block represent the framewise features with the same dimension, we use the same symbol $x$ for simplicity.



\subsection{Voting block architecture}\label{subsec:vote}
The voting block aims to globally propagate key action representations as prior knowledge into targeted local temporal regions. Given input video features $x$, it independently generates two votes for each frame, with one toward the start of the action and the other toward its end. Despite originating from the same frame, these two votes serve different purposes by enhancing features in different action-transiting regions. Unlike conventional Hough voting which computes votes based on a pre-defined lookup table, our approach directly generates votes through the deep network.

More specifically, for video features $x=[x_1, ..., x_T] \in \mathbb{R}^{T\times D}$, where $D$ represents the framewise feature dimension, we first compute the normalized temporal indexes $r \in \mathbb{R}^{T\times 1}$ by dividing frame indexes $[1, 2, ..., T]$ by $T$. This step records the normalized position of each frame in the video. Since each frame outputs two votes targeting different positions, we apply two independent networks, denoted as $start\_net$ and $end\_net$, for the generation of separate vote sets. The $start\_net$ consists of three layers of 1D convolution, with batch normalization and ReLU activation following the first two layers. It takes the video features $x=[x_1, ..., x_T] \in \mathbb{R}^{T\times D}$ as inputs, maintaining the same feature dimensions in the first two 1D convolution layers and modifying the feature dimensions in the outputs of the last layer, which are represented as $[\Delta r_s; \Delta x_s] \in \mathbb{R}^{T\times (1+D)}$, where $\Delta r_s \in \mathbb{R}^{T\times 1}$ denotes the index offsets and $\Delta x_s \in \mathbb{R}^{T\times D}$ represents the feature offsets. The $end\_net$ shares the same architecture as $start\_net$ and follows the same procedure to generate $[\Delta r_e; \Delta x_e] \in \mathbb{R}^{T\times (1+D)}$, where the subscripts $s$ and $e$ represent the outputs from $start\_net$ and $end\_net$ respectively. While the feature offsets provide key action representations as global prior knowledge, the index offsets temporally shift the original features into new positions. Consequently, the voting block enables the model to propagate the action knowledge to any targeted temporal position in the video. Given the original features, indexes, and their corresponding offsets, the vote set can be generated as follows:

\begin{align}
\Delta x_{s}, \Delta r_{s} = start\_net(x), \\
\left\{x'_s, r'_s\right\} = \left\{x + \Delta x_{s}, r + \Delta r_{s}\right\},\\
\Delta x_{e}, \Delta r_{e} = end\_net(x), \\
\left\{x'_e, r'_e\right\}  = \left\{x + \Delta x_{e}, r + \Delta r_{e}\right\},
\end{align}
where $x'_s, x'_e \in \mathbb{R}^{T\times D}$ and $r'_s, r'_e \in \mathbb{R}^{T\times 1}$. We then concatenate $\left\{x'_s, r'_s\right\}$ and $\left\{x'_e, r'_e\right\}$ into vote set $\left\{x', r' \right\}$, where vote features $x' \in \mathbb{R}^{2T\times D}$ and vote temporal indexes $r' \in \mathbb{R}^{2T\times 1}$.

While votes and frames share the same feature dimension, the generated votes exhibit new characteristics in the temporal domain, as the block rearranges the evenly distributed frame features and ensures votes cluster around boundaries. The voting procedure enables our BVN to automatically select global prior knowledge containing rich action representations, which serves as a fundamental basis for suppressing originally less discriminative action-transiting regions. In the next subsection, we will discuss the votes aggregation block for action-transiting feature enhancement.

\subsection{Votes aggregation}\label{subsec:agg}
Based on the votes generated from the voting block, the aggregation block follows a two-step process. It first groups the votes and subsequently aggregates them back into the original video features.

Due to the uneven temporal distribution of votes, selecting representative votes for grouping without introducing noise poses a significant challenge. Inspired by the unstructured spatial distribution observed in 3D point clouds~\cite{zheng2023learning} and the corresponding farthest point sampling technique for subset selection~\cite{qi2017pointnet++}, we introduce temporal farthest point sampling (TFPS) into action segmentation for the first time. TFPS targets $P$ ``key votes'' from the vote set $\left\{x', r' \right\}$ by iteratively selecting points that maximize the minimum distance to already chosen points. In contrast to traditional FPS which calculates distance based on 3D coordinates, we measure the distance between two votes by subtracting their vote temporal indexes $r'$. Subsequently, for each vote in the vote set, we compute its distance to all key votes and select the minimum value, representing its distance to the nearest key vote. If the distance is below the threshold $E$, the vote is included in the group of the corresponding key vote. By repeating the above procedure for all votes, we obtain the final key vote groups $g = {g_1, ..., g_P}$. As votes naturally cluster around action boundaries, the grouping procedure with a distance threshold ensures the selection of sufficiently representative votes for the subsequent aggregation procedure.

Each obtained key vote group conveys essential information from two perspectives. First, it provides key action representations as “what to vote,” represented by the features of all votes within the group. Second, it provides temporal information about “where to vote,” represented by the centroid of the key vote group. Therefore, following the grouping procedure, we pass each key vote group through a shared multi-layer perceptron (MLP) network with max pooling, transforming the key vote group features into a $D$-dimensional feature for aggregation. Subsequently, we retrieve the normalized temporal index associated with the centroid of each key vote group, and aggregate the acquired $D$-dimensional feature into the original video features $x$ with the corresponding temporal index. To elaborate, for each key vote group $g_p$ with centroid's normalized temporal index $c$, the aggregation process is represented as follows:
\begin{align}
g_p' = \text{MP}(\text{MLP}(g_p)), \\
x^{c+i}_{c-i} = x^{c+i}_{c-i} + g_p', \label{eq:agg}
\end{align}
where MP represents max pooling, $g_p'$ denotes the aggregated feature of the key vote group $g_p$, and $x^{c+i}_{c-i}$ indicates the video feature list with central normalized temporal index as $c$ and window size as $2i+1$. Following the aggregation, the updated video features are forwarded to the subsequent decoder layer while retaining the same dimension. We sequentially concatenate multiple stages with the same architecture of voting block, aggregation block, and decoder, and generate the framewise predictions $y^T_1$ from the final decoder using the MLP network.

By proposing voting and aggregation blocks across stages, our BVN demonstrates strong capability in hierarchically propagating and aggregating global prior knowledge from votes into less discriminative regions. Nevertheless, to enable BVN for continuous action-transiting region suppression and boundary localization refinement, two essential prerequisites are required: firstly, identifying action-transiting regions between annotated timestamps, and secondly, precisely targeting generated votes into specified regions. The subsequent subsections will delve into these requirements, beginning with generating action-transiting regions and then designing loss functions for positioning votes.

\subsection{Generating action-transiting regions}\label{subsec:region}
With the provided training video, the prevailing approach for network learning involves generating framewise pseudo-labels based on annotated timestamps $\left\{l^N_1, a^N_1\right\}$, followed by training the segmentation model in a fully-supervised manner. Notably, the boundary localization between each pair of annotated timestamps is equivalent to the pseudo-label generation, as every frame between the annotated timestamp and the obtained boundary can be assigned the same class label based on this annotation. In the following, we will discuss the boundary localization process and the subsequent generation of action-transiting regions.

Consistent with the approach in~\cite{li2021temporal}, we employ bidirectional boundary detection on the video features $x=[x_1, ..., x_T]$, aiming to determine the optimal time, denoted as $\tau$, that partitions the selected period of the video sequence into two clusters. The objective is to minimize the feature distance between every frame and its cluster center. Specifically, utilizing annotated timestamp locations $l_{n}, l_{n+1} \in l^N_1$, we identify the boundary location $b_{n}$ between them as follows:

\begin{align}
b^{FW}_{n} = \underset{\tau}{\arg \min }{\sum_{t=b^{FW}_{n-1}}^{\tau}{d(x_t, m_n)} + \sum_{t=\tau+1}^{l_{n+1}}{d(x_t, m_{n+1})}}, \\
b^{BW}_{n} = \underset{\tau}{\arg \min }{\sum_{t=l_{n}}^{\tau}{d(x_t, m_n)} + \sum_{t=\tau+1}^{b^{BW}_{n+1}}{d(x_t, m_{n+1})}},
\end{align}
where $FW$, $BW$ represents forward and backward, while $d(., .)$ calculates the Euclidean distance between the given feature vectors. $m$ denotes the mean of all features $x_{t}$ within the upper-lower bound of summation. Therefore, $m_{n}$ and $m_{n+1}$ represent the mean feature of $x_{t}$ within the first and second summations, as indicated in the formulas.

We apply forward and backward boundary detection separately on video features $x$ to generate two sets of boundary estimations: $b^{FW}$ and $b^{BW}$, resulting in two sets of framewise pseudo-labels. Due to the inherent feature ambiguity, these two sets of generated framewise pseudo-labels may conflict with frames around the boundaries. In such cases, the action-transiting region between the $n^{th}$ and ${(n+1)}^{th}$ actions can be represented as:

\begin{align}
t \in [\min(b^{FW}_{n}, b^{BW}_{n}), \max(b^{FW}_{n}, b^{BW}_{n})]. \label{eq:region}
\end{align}

By iteratively applying the voting blocks and aggregation blocks on the video features, the temporal duration of action-transiting regions undergoes continuous suppression. This demonstrates that our BVN effectively mitigates feature ambiguity and uncertainty in boundary localization. In the experiment section, we provide detailed visualizations to further illustrate the reduction in action-transiting regions across different stages.

\subsection{Loss functions and training process}\label{subsec:loss}
Existing timestamp-supervised action segmentation approaches~\cite{li2021temporal, zhao2022turning, liu2023reducing} commonly incorporate framewise classification loss, smoothing loss, and confidence loss during the training phase. In addition to these three, we introduce the voting loss to ensure that each vote is directed towards the corresponding boundary neighboring regions. We provide further details on the losses below.

\textbf{Framewise classification loss.} We apply the cross-entropy loss to the model predictions. Here $y_{t, \tilde{a}}$ represents the predicted probability of ground truth action $\tilde{a}$ for time $t$.

\begin{align}
\mathcal{L}_{frame} = \frac{1}{T} \sum_{t=1}^{T}-\log(y_{t, \tilde{a}}).
\end{align}

\textbf{Smoothing loss.} To alleviate the over-segmentation problem, we calculate the truncated mean square error~\cite{farha2019ms}:

\begin{align}
\mathcal{L}_{smooth} = \frac{1}{TA} \sum_{t=1}^{T} \sum_{a=1}^{A} {\min(|\log{y_{t, a}} - \log{y_{t-1, a}}|, \delta)}^2,
\end{align}
where $A$ represents the total number of action classes, $y_{t, a}$ indicates the predicted probability of action $a$ for time $t$, and $\delta$ is a constant hyperparameter. The loss enhances consistency between neighboring frames and ensures a smooth transition in predictions.

\textbf{Confidence loss.} With frames distant from the annotated timestamp $l_n$, confidence in predicting them as $a_n$ decreases. The confidence loss is designed accordingly:

\begin{align}
\mathcal{L}_{conf} = \frac{1}{2(l_N - l_1)} \sum_{n=1}^{N}{\Big(\sum_{t=l_{n-1}}^{l_{n+1}}\theta_{a_{l_n}, t}\Big)},
\end{align}

\begin{equation}\label{eq2}
	\theta_{a_{l_n}, t}=\left\{
	\begin{aligned}
		\max(0, \log{y_{t-1, a_{l_n}}} - \log{y_{t, a_{l_n}}}) & , & t < l_n\\
		\max(0, \log{y_{t, a_{l_n}}} - \log{y_{t-1, a_{l_n}}}) & , & t \ge l_n
	\end{aligned}
	\right. ,
\end{equation}
where $y_{t, a_{l_n}}$ represents the predicted probability of action $a_{l_n}$ for time $t$.

\begin{table*}
    \caption{Action segmentation performance on the GTEA, 50Salads, and Breakfast datasets, with the table organized according to different backbones. The dash (-) symbol indicates that no prior result is available.}
    \label{tab:res1}
    \centering
    \scalebox{1.05}{
    \begin{tabular}{l|c|c|c|c|c|c|c|c|c}
        \hline
         & \multicolumn{3}{c|}{\textbf{GTEA}} & \multicolumn{3}{c|}{\textbf{50Salads}} & \multicolumn{3}{c}{\textbf{Breakfast}} \\
        \hline
        \textbf{Methods} & F1@\{10, 25, 50\} & Edit & Acc & F1@\{10, 25, 50\} & Edit & Acc & F1@\{10, 25, 50\} & Edit & Acc \\
        \hline
        Li et al.~\cite{li2021temporal} & 78.9 \hspace{0.5em} 73.0 \hspace{0.5em} 55.4 & 72.3 & 66.4 & 73.9 \hspace{0.5em} 70.9 \hspace{0.5em} 60.1 & 66.8 & 75.6 & 70.5 \hspace{0.5em} 63.6 \hspace{0.5em} 47.4 & 69.9 & 64.1 \\
        Zhao et al.~\cite{zhao2022turning} & 84.3 \hspace{0.5em} 81.7 \hspace{0.5em} 64.8 & 79.8 & 74.4 & 78.5 \hspace{0.5em} 75.5 \hspace{0.5em} 63.4 & 71.8 & 77.7 & 73.1 \hspace{0.5em} 66.5 \hspace{0.5em} 49.4 & 72.6 & 64.6 \\
        Khan et al.~\cite{khan2022timestamp} & 81.5 \hspace{0.5em} 77.5 \hspace{0.5em} 60.8 & 75.6 & 66.1 & 75.1 \hspace{0.5em} 72.3 \hspace{0.5em} 61.0 & 67.6 & 75.1 & 67.9 \hspace{0.5em} 61.0 \hspace{0.5em} 45.3 & 67.0 & 61.4 \\
        EM-TSS~\cite{rahaman2022generalized} & \hspace{0.7em}- \hspace{1.3em} 82.7 \hspace{0.5em} 66.5 & 82.3 & 70.5 & \hspace{0.7em}- \hspace{1.3em} 75.9 \hspace{0.5em} 64.7 & 71.6 & 77.9 & \hspace{0.7em}- \hspace{1.3em} 63.7 \hspace{0.5em} 49.8 & 67.2 & 67.0 \\
        Du et al.~\cite{du2022timestamp} & 83.7 \hspace{0.5em} 79.8 \hspace{0.5em} 65.4 & 77.2 & 70.1 & 77.3 \hspace{0.5em} 74.7 \hspace{0.5em} 63.7 & 70.1 & 78.6 & 71.2 \hspace{0.5em} 64.6 \hspace{0.5em} 48.9 & 71.6 & 65.7 \\
        RWS~\cite{hirsch2024random} & 80.9 \hspace{0.5em} 74.1 \hspace{0.5em} 56.3 & 76.2 & 59.3 & 76.7 \hspace{0.5em} 72.8 \hspace{0.5em} 55.5 & 69.3 & 70.0 & 70.9 \hspace{0.5em} 64.7 \hspace{0.5em} 44.8 & 71.1 & 60.2 \\
        Sayed et al.~\cite{sayed2023new} & 82.1 \hspace{0.5em} 78.7 \hspace{0.5em} 63.0 & 74.8 & 70.4 & 77.3 \hspace{0.5em} 75.2 \hspace{0.5em} 63.6 & 69.8 & 75.8 & - \hspace{0.5em} - \hspace{0.5em} - & - & - \\
        TSCL~\cite{patsch2024tscl} & \textbf{88.5} \hspace{0.5em} \textbf{84.9} \hspace{0.5em} \textbf{69.0} & 83.0 & 74.5 & 79.5 \hspace{0.5em} 76.2 \hspace{0.5em} 61.8 & \textbf{73.7} & 74.9 & 71.3 \hspace{0.5em} 64.3 \hspace{0.5em} 47.5 & 68.9 & 64.8 \\
        \textbf{BVN (MS-TCN++)} & 86.2 \hspace{0.5em} \textbf{84.9} \hspace{0.5em} 68.2 & \textbf{84.7} & \textbf{75.5} & \textbf{81.4} \hspace{0.5em} \textbf{76.8} \hspace{0.5em} \textbf{67.8} & 73.5 & \textbf{80.1} & \textbf{74.4} \hspace{0.5em} \textbf{68.0} \hspace{0.5em} \textbf{50.4} & \textbf{74.0} & \textbf{67.8} \\
        \hline
        UVAST + alignment~\cite{behrmann2022unified} & 70.8 \hspace{0.5em} 63.5 \hspace{0.5em} 49.2 & 88.2 & 55.3 & 75.7 \hspace{0.5em} 70.6 \hspace{0.5em} 58.2 & 78.4 & 67.8 & 72.0 \hspace{0.5em} 64.1 \hspace{0.5em} 48.6 & 74.3 & 60.2 \\
        UVAST + Viterbi~\cite{behrmann2022unified} & 87.2 \hspace{0.5em} 83.7 \hspace{0.5em} 66.0 & 89.3 & 70.5 & 83.0 \hspace{0.5em} 79.6 \hspace{0.5em} 65.9 & 78.2 & 77.0 & 71.3 \hspace{0.5em} 63.3 \hspace{0.5em} 48.3 & 74.1 & 60.7 \\
        UVAST + FIFA~\cite{behrmann2022unified} & 80.7 \hspace{0.5em} 75.2 \hspace{0.5em} 57.4 & 88.7 & 66.0 & 80.2 \hspace{0.5em} 74.9 \hspace{0.5em} 61.6 & 78.6 & 72.5 & 72.0 \hspace{0.5em} 64.2 \hspace{0.5em} 47.6 & 74.1 & 60.3 \\
        Yang et al.~\cite{yang2023weakly}       & 88.2 \hspace{0.5em} 85.5 \hspace{0.5em} 67.3 & 84.0 & 69.2 & 84.4 \hspace{0.5em} 81.3 \hspace{0.5em} 67.8 & 77.9 & 77.0 & 71.5 \hspace{0.5em} 64.2 \hspace{0.5em} 47.0 & 72.3 & 64.6 \\
        D-TSTAS~\cite{liu2023reducing} & 91.5 \hspace{0.5em} 90.1 \hspace{0.5em} 76.2 & 88.5 & 75.7 & 84.2 \hspace{0.5em} 82.1 \hspace{0.5em} 71.5 & 77.6 & 80.0 & 76.7 \hspace{0.5em} 69.3 \hspace{0.5em} 50.7 & 75.8 & 65.7 \\
        \textbf{BVN (ASFormer)} & \textbf{91.7} \hspace{0.5em} \textbf{90.5} \hspace{0.5em} \textbf{76.8} & \textbf{89.9} & \textbf{76.7} & \textbf{85.1} \hspace{0.5em} \textbf{82.5} \hspace{0.5em} \textbf{72.3} & \textbf{79.0} & \textbf{81.2} & \textbf{77.3} \hspace{0.5em} \textbf{69.7} \hspace{0.5em} \textbf{51.2} & \textbf{76.5} & \textbf{68.3} \\
        \hline
    \end{tabular}}
\end{table*}

\begin{table*}
    \caption{Framewise pseudo-label generation performance on the GTEA, 50Salads, and Breakfast datasets, with the table organized according to different backbones.}
    \label{tab:pseudo}
    \centering
    \scalebox{1.1}{
    \begin{tabular}{l|c|c|c|c|c|c}
        \hline
         & \multicolumn{2}{c|}{\textbf{GTEA}} & \multicolumn{2}{c|}{\textbf{50Salads}} & \multicolumn{2}{c}{\textbf{Breakfast}} \\
        \hline
        \textbf{Methods} & F1@\{10, 25, 50\} & Acc & F1@\{10, 25, 50\} & Acc & F1@\{10, 25, 50\} & Acc \\
        \hline
        Li et al.~\cite{li2021temporal} & 96.6 \hspace{0.5em} 86.3 \hspace{0.5em} 65.5 & 75.5 & 99.4 \hspace{0.5em} 95.0 \hspace{0.5em} 76.8 & 79.9 & 96.0 \hspace{0.5em} 87.3 \hspace{0.5em} 67.6 & 72.6 \\
        RWS~\cite{hirsch2024random} & 96.7 \hspace{0.5em} 89.4 \hspace{0.5em} 71.4 & 78.6 & 99.7 \hspace{0.5em} 97.5 \hspace{0.5em} 81.0 & 80.6 & 96.5 \hspace{0.5em} 88.9 \hspace{0.5em} 69.5 & 76.1 \\
        \textbf{BVN (MS-TCN++)} & \textbf{98.6} \hspace{0.5em} \textbf{95.3} \hspace{0.5em} \textbf{78.9} & \textbf{79.4} & \textbf{99.8} \hspace{0.5em} \textbf{97.9} \hspace{0.5em} \textbf{82.5} & \textbf{82.2} & \textbf{97.1} \hspace{0.5em} \textbf{90.0} \hspace{0.5em} \textbf{71.7} & \textbf{77.9} \\
        \hline
        UVAST~\cite{behrmann2022unified} & \textbf{99.8} \hspace{0.5em} 97.7 \hspace{0.5em} 83.0 & 75.3 & 97.5 \hspace{0.5em} 90.4 \hspace{0.5em} 75.6 & 81.3 & 95.5 \hspace{0.5em} 87.5 \hspace{0.5em} 70.0 & 76.9 \\
        \textbf{BVN (ASFormer)} & \textbf{99.8} \hspace{0.5em} \textbf{98.2} \hspace{0.5em} \textbf{85.9} & \textbf{80.5} & \textbf{99.9} \hspace{0.5em} \textbf{98.3} \hspace{0.5em} \textbf{84.7} & \textbf{83.4} & \textbf{97.8} \hspace{0.5em} \textbf{90.7} \hspace{0.5em} \textbf{73.4} & \textbf{79.8} \\
        \hline
    \end{tabular}}
\end{table*}

\textbf{Voting loss.} We propose the voting loss to ensure that each vote is directed toward the corresponding action-transiting region. Specifically, our $start\_net$ and $end\_net$ independently guide votes from each frame into their respective starting and ending action-transiting regions. For the region between the $n^{th}$ and ${(n+1)}^{th}$ actions, as defined in Eq.~\eqref{eq:region}, the center is computed as $\frac{b^{FW}{n} + b^{BW}{n}}{2}$. For frames belonging to the ${(n+1)}^{th}$ action, we determine the ground truth index offsets for $start\_net$ by subtracting the index of each frame from the center. We follow the same process for $end\_net$ and supervise the temporal offsets learning using the regression loss:

\begin{align}
\mathcal{L}_{vote} = \frac{1}{2T} \sum_{t=1}^{T}((\Delta r^*_{s} - \Delta r_{s})^2 + (\Delta r^*_{e} - \Delta r_{e})^2),
\end{align}
where $\Delta r^*_{s}$ and $\Delta r^*_{e}$ represent the ground truth index offsets of the $start\_net$ and $end\_net$. $\Delta r_{s}$ and $\Delta r_{e}$ denote the predicted index offsets from these two networks, as detailed in Sec.~\ref{subsec:vote}.

In summary, the final loss function for training the action segmentation model is expressed as:

\begin{align}
\mathcal{L} = \mathcal{L}_{frame} + \alpha\mathcal{L}_{smooth} + \beta\mathcal{L}_{conf} + \gamma\mathcal{L}_{vote},
\end{align}
where different losses are balanced using the hyperparameters $\alpha$, $\beta$, and $\gamma$.

We adopt the same two-stage training pipeline as described in~\cite{li2021temporal}, consisting of the initialization training and the iterative training. The initialization stage utilizes only the annotated timestamps for training, which provides fundamental knowledge for network learning. For the iterative training stage, the network is supervised using the generated framewise pseudo-labels, while iteratively refining the quality of these pseudo-labels as training progresses. The voting loss is introduced after the initialization stage.

\subsection{Discussions with related TCSVT papers}
Zou \textit{et al}.\cite{zou2024weakly} and Li \textit{et al}.\cite{hao2024hierarchical} adopted transcript supervision and dense multi-task annotations, respectively, whereas our paper employs timestamp supervision. Li \textit{et al}.\cite{li2023involving} applied a graph-based convolutional network for human skeleton joint data, while we focus on addressing feature-level ambiguity in conventional RGB-based features. Du \textit{et al}.\cite{du2024weakly} and Shao \textit{et al}.~\cite{shao2024text} formulated the problem as predicting each action segment’s category, confidence score, start time, and end time, whereas we adhere to the classical action segmentation paradigm by generating framewise action predictions. Compared to the aforementioned papers, we propose the first work to address feature-level ambiguity in action-transiting regions through a global-to-local voting mechanism, distinguishing itself in supervision format, methodological framework, and problem formulation.

\begin{figure*}[t]
 \centering
 \includegraphics[width=1.0\linewidth]{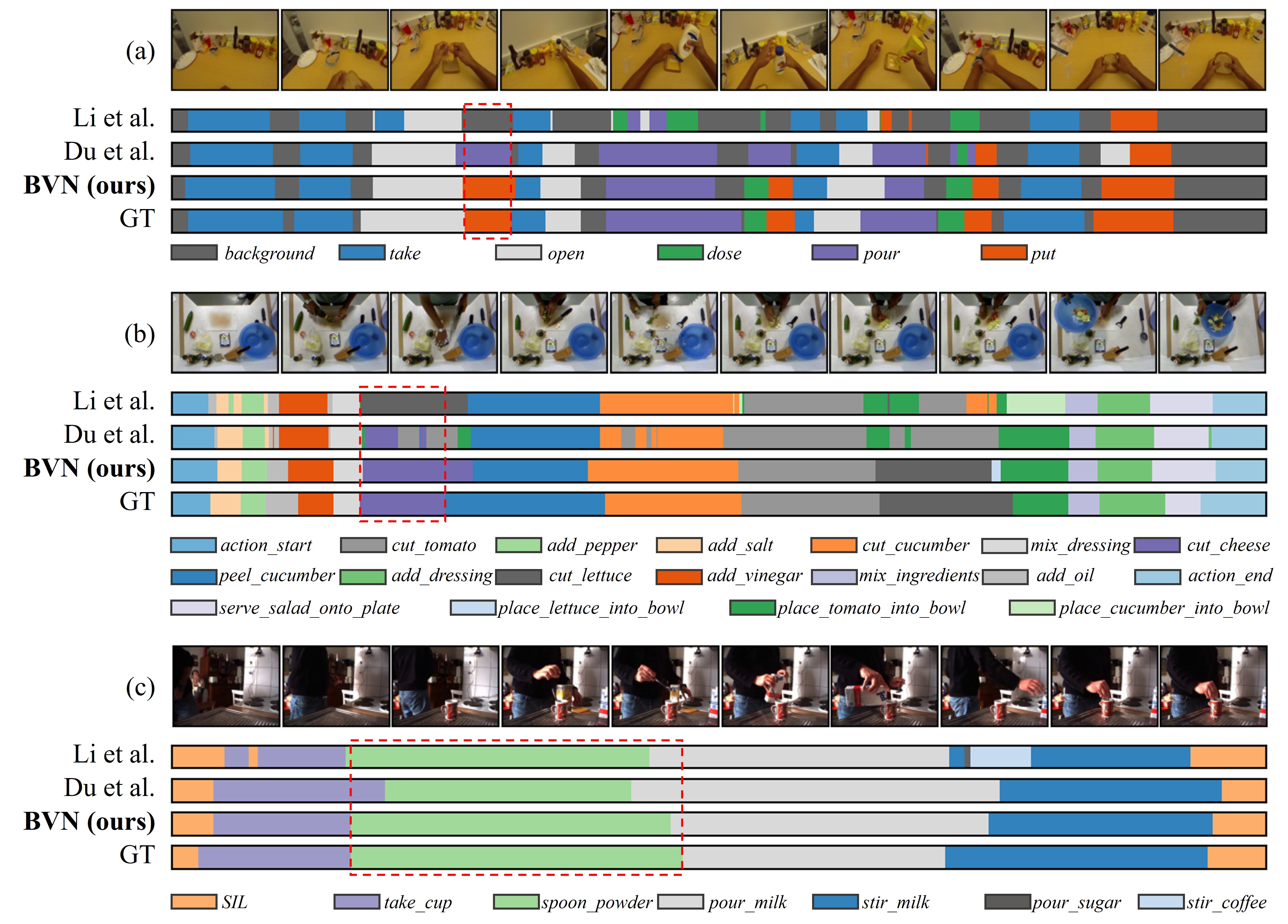}
  \caption{Action segmentation visualizations on the (a) GTEA, (b) 50Salads, and (c) Breakfast video samples, with ``GT'' denoting the ground truth segmentation. Compared to existing methods, our approach demonstrates reliable performance in accurately predicting actions and achieving better alignment with the start and end times of ground truth actions, as highlighted by the red dashed rectangles.}
  \label{fig:visual}
\end{figure*}

\section{Experiments}
\label{sec:blind}
In this section, we provide a brief overview of the experimental setup, followed by both quantitative and qualitative comparisons. We then provide additional quantitative analysis and conduct detailed ablation studies to assess the effectiveness of the proposed method.

\subsection{Experimental setup}
\textbf{Datasets.} We conduct experiments on three real-world video datasets: Georgia Tech Egocentric Activities (GTEA)~\cite{fathi2011learning}, 50Salads~\cite{stein2013combining}, and Breakfast~\cite{kuehne2014language}. The GTEA dataset consists of 58 egocentric instructional videos with 11 action classes, capturing daily activities with an average duration of 1 minute per video. The 50Salads dataset contains 50 instructional videos with 19 actions related to salad preparation, with an average video duration of 6.4 minutes. The Breakfast dataset includes 1712 third-person videos varying from seconds to a few minutes, covering 48 different actions across 10 breakfast-related activities, such as making tea and preparing coffee. To ensure a fair comparison, we employ the same annotations as described in~\cite{li2021temporal}.

\textbf{Evaluation metrics.} We apply the following metrics for evaluation: (1) Framewise accuracy (Acc), which measures the proportion of correctly predicted frames relative to the total number of frames. (2) Segmental edit score (Edit), which quantifies the similarity of the predicted sequence and the ground truth using Levenshtein distance. It evaluates the quality of the action sequence without relying on framewise prediction. (3) The F1 score, which compares the intersection over union (IoU) of each segment with the ground truth. We calculate the scores under three thresholds: F1@\{10, 25, 50\}.

\begin{figure*}[t]
 \centering
 \includegraphics[width=1.0\linewidth]{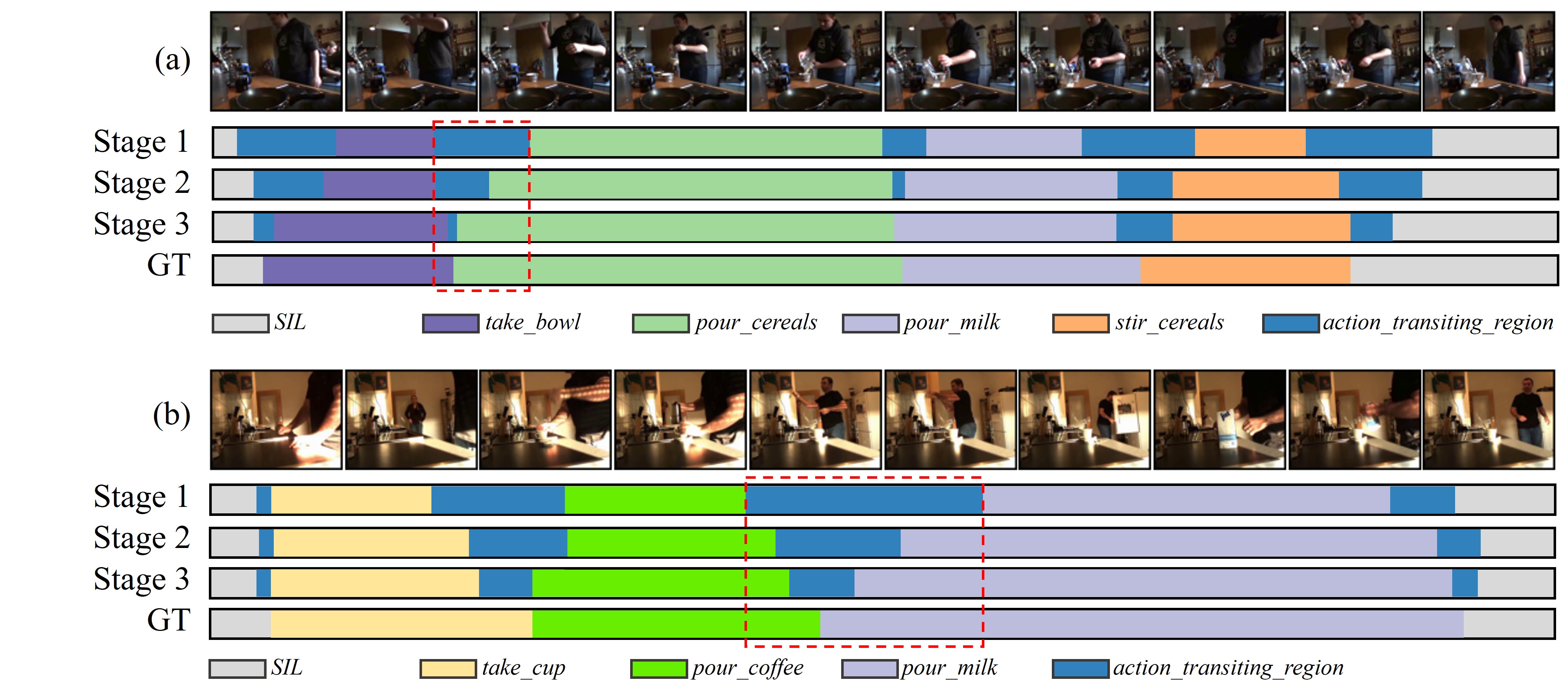}
  \caption{Hierarchical boundary localization refinement during training on the Breakfast video samples, with ``GT” denoting the ground truth segmentation. The action-transiting regions, depicted in blue, are progressively suppressed as the voting mechanism is iteratively applied at different stages, as highlighted by the red dashed rectangles.}
  \label{fig:refine}
\end{figure*}

\begin{figure}[ht]
 \centering
 \includegraphics[width=1.0\linewidth]{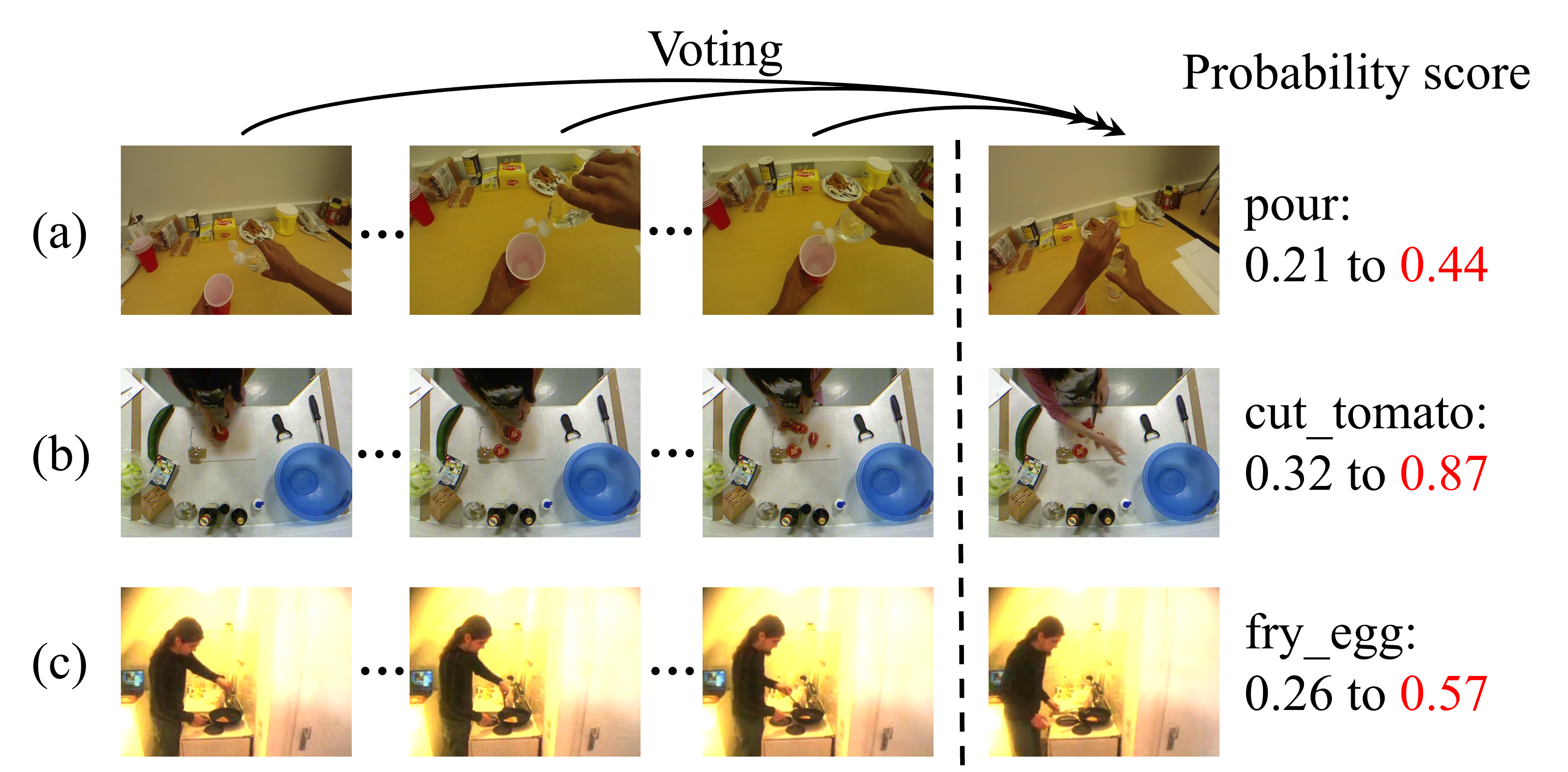}
  \caption{Visualizations of voting from explicit to ambiguous action frames in action-transiting regions, along with the probability score improvements on the (a) GTEA, (b) 50Salads, and (c) Breakfast video samples.}
  \label{fig:ab_voting}
\end{figure}

\textbf{Implementation details.} We utilize pre-extracted I3D~\cite{carreira2017quo} features as the network inputs. For the segmentation model, we employ the same encoder and decoder architecture as in ASFormer~\cite{yi2021asformer} and set the number of decoders to 3. The number of voting and aggregation blocks is the same as the decoders, as these blocks are integrated before each decoder layer. We set the intermediary feature dimension $D$ to 64, the distance threshold $E$ to 0.1, and $i$ in Eq.~\eqref{eq:agg} to 5. We apply random initialization in TFPS to select the first point, and BVN returns stable outputs due to the substantial number of votes and their clustering characteristics around boundaries. During training, the network is initially trained with only annotated timestamps for 70 epochs, followed by training on framewise pseudo-labels for 50 epochs. We set the hyperparameters $\alpha = 0.15$, $\beta = 0.075$, and $\gamma = 0.01$. All experiments are conducted on a single V100 GPU.

\subsection{Quantitative comparisons}
We quantitatively compare the proposed method with existing works from two perspectives: action segmentation performance and framewise pseudo-label generation performance.

\textbf{Action segmentation performance.} Table~\ref{tab:res1} compares the action segmentation performance against previous approaches, with the best results highlighted in bold. Since most existing timestamp-supervised methods utilize MS-TCN++~\cite{li2021temporal, zhao2022turning, khan2022timestamp, rahaman2022generalized, du2022timestamp, hirsch2024random} and ASFormer~\cite{behrmann2022unified, liu2023reducing} as network backbones, we implement the boundary voting network using both architectures for a fair comparison. The table is divided into two sections based on their respective backbones.

By leveraging the voting mechanism to address action-transiting feature ambiguity and refine boundary localization, BVN achieves state-of-the-art results across three datasets under various evaluation metrics. For methods using the MS-TCN++ backbone, our approach demonstrates consistent performance improvements in framewise accuracy: +1.0\% on the GTEA dataset, +1.4\% on the 50Salads dataset, and +0.8\% on the Breakfast dataset. With the ASFormer backbone, our method enhances framewise accuracy by +1.0\% on the GTEA dataset, +1.2\% on the 50Salads dataset, and +2.6\% on the Breakfast dataset. In summary, our proposed model not only consistently outperforms other approaches but also exhibits robust performance with different network backbones.

\textbf{Framewise pseudo-label generation performance.} Precise boundary localization between annotated timestamps is crucial, as it directly influences the quality of the generated framewise pseudo-labels. To demonstrate the effectiveness of boundary localization during training, we present the framewise pseudo-label generation performance in Table~\ref{tab:pseudo} and compare it with the results reported by existing works. Following Table~\ref{tab:res1}, we divide Table~\ref{tab:pseudo} into two sections based on their respective backbones.

Without bells and whistles, our method significantly improves framewise pseudo-label generation results across all datasets, evaluation metrics, and backbones. For example, under the F1@50 metric, our method achieves a +7.5\% improvement on the GTEA dataset with the MS-TCN++ backbone, and a +9.1\% improvement on the 50Salads dataset with the ASFormer backbone. In summary, by reducing boundary localization uncertainty through the voting mechanism, we generate significantly higher-quality framewise pseudo-labels, which, in turn, enhance training stability and improve action segmentation testing performance in Table~\ref{tab:res1}.

\begin{table}
    \centering
    \caption{Action-transiting features quality results on the GTEA, 50Salads, and Breakfast datasets.}
    \scalebox{1.1}{
    \begin{tabular}{l|c|c|c}
        \hline
        \textbf{Methods} & \textbf{GTEA} & \textbf{50Salads} & \textbf{Breakfast} \\
        \hline
        Li et al.~\cite{li2021temporal} &0.69 &0.75 &0.67 \\
        Du et al.~\cite{du2022timestamp} &0.72 &0.79 &0.74 \\
        BVN (MS-TCN++) &0.79 &0.86 &0.81 \\
        \textbf{BVN (ASFormer)} &\textbf{0.82} &\textbf{0.92} &\textbf{0.86} \\
        \hline
    \end{tabular}}
    \label{tab:exp_quality}
\end{table}

\begin{table}
    \centering
    \caption{Index offsets quality results on the GTEA, 50Salads, and Breakfast datasets.}
    \scalebox{1.1}{
    \begin{tabular}{l|c|c|c}
        \hline
        \textbf{Methods} & \textbf{GTEA} & \textbf{50Salads} & \textbf{Breakfast} \\
        \hline
        BVN (MS-TCN++) &0.07 &0.06 &0.04 \\
        BVN (ASFormer) &0.05 &0.05 &0.03 \\
        \hline
    \end{tabular}}
    \label{tab:exp_precison}
\end{table}

\begin{table}[t]
\centering
\caption{Action segmentation performance comparisons with fully-supervised approaches on the 50Salads dataset.}
\scalebox{1.0}{
\begin{tabular}{c|c|c|c|c}
        \hline
        & \textbf{Methods} & \textbf{F1@\{10, 25, 50\}} & \textbf{Edit} & \textbf{Acc} \\
        \hline
        \multirow{4}{*}{\textbf{Original}} 
        & MS-TCN++~\cite{li2020ms}          & 80.7 \hspace{0.5em} 78.5 \hspace{0.5em} 70.1 & 74.3 & 83.7 \\
        & ASFormer~\cite{yi2021asformer}    & 85.1 \hspace{0.5em} 83.4 \hspace{0.5em} 76.0 & 76.9 & 85.6 \\
        & ASPnet~\cite{van2023aspnet}       & 92.7 \hspace{0.5em} 91.6 \hspace{0.5em} 88.5 & 87.5 & 91.4 \\
        & BaFormer~\cite{wang2024efficient} & 89.3 \hspace{0.5em} 88.4 \hspace{0.5em} 83.9 & 84.2 & 89.5 \\
        \hline
        \multirow{4}{*}{\textbf{50\%}} 
        & MS-TCN++~\cite{li2020ms}          & 70.5 \hspace{0.5em} 64.4 \hspace{0.5em} 55.6 & 61.6 & 66.5 \\
        & ASFormer~\cite{yi2021asformer}    & 73.6 \hspace{0.5em} 67.9 \hspace{0.5em} 61.3 & 68.3 & 71.4 \\
        & ASPnet~\cite{van2023aspnet}       & 76.6 \hspace{0.5em} 68.2 \hspace{0.5em} 60.3 & 68.7 & 70.7 \\
        & BaFormer~\cite{wang2024efficient} & 79.5 \hspace{0.5em} 74.3 \hspace{0.5em} 65.5 & 72.7 & 75.5 \\
        \hline
        \multirow{1}{*}{\textbf{Timestamp}} 
        & BVN (Ours)                        & 85.1 \hspace{0.5em} 82.5 \hspace{0.5em} 72.3 & 79.0 & 81.2 \\
        \hline
    \end{tabular}}
    \label{tab:vs_fully}
\end{table}

\begin{table}[t]
    \centering
    \caption{Comparisons of computational complexity.}
    \scalebox{1.0}{
    \begin{tabular}{l|c|c|c}
        \hline
        \textbf{Methods} & \textbf{Parameter num} & \textbf{FLOPs} & \textbf{GPU usage} \\
        \hline
        Li et al.~\cite{li2021temporal} &1.2M &2.2G &7.6G \\
        BVN (MS-TCN++) &1.4M &2.6G &9.9G \\
        BVN (ASFormer) &1.9M &3.7G &15.5G \\
        \hline
    \end{tabular}}
    \label{tab:exp_computational}
\end{table}

\subsection{Qualitative comparisons}
We qualitatively compare the proposed method with existing works from two perspectives: action segmentation visualizations and boundary localization refinement visualizations. We further visualize the voting procedure for ambiguous action frames and the corresponding probability score improvements.

\textbf{Action segmentation visualizations.} Fig.~\ref{fig:visual} presents action segmentation visualizations from the methods of Li \textit{et al}.~\cite{li2021temporal}, Du \textit{et al}.~\cite{du2022timestamp}, and our proposed BVN model, where distinct colors denote different action segments. Compared to existing methods, our method demonstrates superior segmentation performance by effectively capturing challenging actions missed by other methods and providing more precise boundary localization, particularly in long videos with complex actions.

\textbf{Boundary localization refinement visualizations.} 
We visualize the boundary localization refinement for framewise pseudo-label generation during training. In Fig.~\ref{fig:refine}, we represent the sequence of the voting block, aggregation block, and decoder as a stage, and generate action-transiting regions on top of the video feature outputs from each stage. For simplicity and clarity, we select the video samples from the Breakfast dataset, which contains only a limited number of action segments. By applying the voting mechanism across the video, the action-transiting features integrate rich contextual knowledge and exhibit more discriminative representations, progressively suppressing the action-transiting regions and refining the boundary localization decisions.

\textbf{Voting visualizations.} 
To assess the concrete influence of voting mechanism, we visualize the voting procedure for ambiguous action frames and the corresponding probability score improvements. As shown in Fig.~\ref{fig:ab_voting}, our voting mechanism gathers essential global prior knowledge from frames depicting explicit ``pour," ``cut\_tomato," and ``fry\_egg" actions and accurately directs it toward ambiguous frames in action-transiting regions. Additionally, compared to the baseline without voting, aggregation module, and voting loss, BVN significantly improves the probability scores of frames with ambiguous visual appearance, demonstrating the effectiveness of voting in mitigating feature ambiguity.

\subsection{Additional quantitative analysis}
In addition to the quantitative comparisons presented in Table~\ref{tab:res1} and~\ref{tab:pseudo}, we provide further analysis from four key perspectives: the quality of action-transiting features, the quality of index offsets, comparisons with fully-supervised setting,  and computational complexity. The additional analysis offer deeper insights into the voting mechanism and its contributions to the overall action segmentation performance.

\textbf{Action-transiting features quality.} As the first work analyzing the quality of action-transiting features, we propose the following evaluation pipeline. For each testing video and its ground truth labels, we identify the center frame of each ground truth action as the ``key action frame" and collect 10 adjacent frames near the ground truth boundary as ``ambiguous frames". Using the trained network, we compute the framewise feature outputs from the final decoder and calculate the average cosine similarity between the ``key action frame" and the ``ambiguous frames". As shown in Table~\ref{tab:exp_quality}, our method significantly improves the feature similarity compared to existing methods, demonstrating the effectiveness of action-transiting feature enhancement.

\textbf{Index offsets quality.} To analyze the quality of the index offsets, we propose the following evaluation pipeline. For each testing video and its ground truth labels, we select the votes generated in the final voting block and calculate the average temporal distance between each vote and its corresponding ground truth boundary. As shown in Table~\ref{tab:exp_precison}, an average error of 0.03 on the Breakfast dataset for the ASFormer backbone indicates that for a 1000-frame video, the index error is approximately 30 frames from the boundary, suggesting that the majority of votes are concentrated around the action-transiting regions. Nevertheless, as we are the first to address timestamp-supervised action segmentation from a voting perspective, no prior works are available for direct comparison.


\begin{table*}
    \caption{Action segmentation performance compared with the baseline on the GTEA, 50Salads, and Breakfast datasets.}
    \label{tab:ab_baseline}
    \centering
    \scalebox{1.1}{
    \begin{tabular}{l|c|c|c|c|c|c|c|c|c}
        \hline
         & \multicolumn{3}{c|}{\textbf{GTEA}} & \multicolumn{3}{c|}{\textbf{50Salads}} & \multicolumn{3}{c}{\textbf{Breakfast}} \\
        \hline
         & \textbf{F1@\{10, 25, 50\}} & \textbf{Edit} & \textbf{Acc} & \textbf{F1@\{10, 25, 50\}} & \textbf{Edit} & \textbf{Acc} & \textbf{F1@\{10, 25, 50\}} & \textbf{Edit} & \textbf{Acc} \\
        \hline
        Baseline & 83.8 \hspace{0.5em} 81.4 \hspace{0.5em} 63.2 & 81.9 & 73.1 & 79.9 \hspace{0.5em} 74.8 \hspace{0.5em} 64.6 & 71.4 & 78.1 & 72.9 \hspace{0.5em} 66.0 \hspace{0.5em} 48.8 & 72.3 & 64.4 \\
        \textbf{BVN (MS-TCN++)} & \textbf{86.2} \hspace{0.5em} \textbf{84.9} \hspace{0.5em} \textbf{68.2} & \textbf{84.7} & \textbf{75.5} & \textbf{81.4} \hspace{0.5em} \textbf{76.8} \hspace{0.5em} \textbf{67.8} & \textbf{73.5} & \textbf{80.1} & \textbf{74.4} \hspace{0.5em} \textbf{68.0} \hspace{0.5em} \textbf{50.4} & \textbf{74.0} & \textbf{67.8} \\
        \hline
        Baseline & 86.6 \hspace{0.5em} 83.0 \hspace{0.5em} 72.3 & 84.3 & 72.7 & 82.1 \hspace{0.5em} 77.8 \hspace{0.5em} 67.8 & 75.3 & 78.2 & 73.0 \hspace{0.5em} 65.6 \hspace{0.5em} 48.0 & 73.9 & 63.8 \\
        \textbf{BVN (ASFormer)} & \textbf{91.7} \hspace{0.5em} \textbf{90.5} \hspace{0.5em} \textbf{76.8} & \textbf{89.9} & \textbf{76.7} & \textbf{85.1} \hspace{0.5em} \textbf{82.5} \hspace{0.5em} \textbf{72.3} & \textbf{79.0} & \textbf{81.2} & \textbf{77.3} \hspace{0.5em} \textbf{69.7} \hspace{0.5em} \textbf{51.2} & \textbf{76.5} & \textbf{68.3} \\
        \hline
    \end{tabular}}
\end{table*}

\begin{table}[tbp!]
\centering
\caption{Framewise pseudo-label generation performance compared with the baseline on the GTEA dataset.}
\scalebox{1.1}{
\begin{tabular}{c|c|c}
        \hline
         & \textbf{F1@\{10, 25, 50\}} & \textbf{Acc}\\
       \hline
        Baseline         & 97.0 \hspace{0.5em} 94.9 \hspace{0.5em} 77.0 & 77.6 \\
        \textbf{BVN (ASFormer)}    & \textbf{99.8} \hspace{0.5em} \textbf{98.2} \hspace{0.5em} \textbf{85.9} & \textbf{80.5} \\
        \hline
    \end{tabular}}
    \label{tab:ab_global_prior}
\end{table}

\begin{table}[tbp!]
\centering
\caption{Effect of the proposed voting and aggregation blocks on the GTEA dataset.}
\scalebox{0.9}{
\begin{tabular}{c|c|c|c}
        \hline
        \textbf{MS-TCN++} & \textbf{F1@\{10, 25, 50\}} & \textbf{Edit} & \textbf{Acc}\\
       \hline
        Voting block only                           &83.0 \hspace{0.5em} 81.0 \hspace{0.5em} 62.7 & 80.6 & 72.4 \\
        Voting + Feature max pooling                 &83.9 \hspace{0.5em} 82.8 \hspace{0.5em} 65.3 & 82.4 & 73.6 \\
        Voting + Aggregation (average pooling)       &85.7 \hspace{0.5em} 84.5 \hspace{0.5em} 67.3 & 84.3 & 75.2 \\
        \textbf{Voting + Aggregation (max pooling)}  &\textbf{86.2} \hspace{0.5em} \textbf{84.9} \hspace{0.5em} \textbf{68.2} & \textbf{84.7} & \textbf{75.5} \\
        \hline
        \textbf{ASFormer} & \textbf{F1@\{10, 25, 50\}} & \textbf{Edit} & \textbf{Acc}\\
       \hline
        Voting block only                           &87.3 \hspace{0.5em} 81.9 \hspace{0.5em} 71.8 &84.9 & 72.2 \\
        Voting + Feature max pooling                 &88.3 \hspace{0.5em} 85.9 \hspace{0.5em} 73.7 &85.2 &74.0 \\
        Voting + Aggregation (average pooling)       &91.5 \hspace{0.5em} 89.4 \hspace{0.5em} 76.1 &89.8 &75.7 \\
        \textbf{Voting + Aggregation (max pooling)}  &\textbf{91.7} \hspace{0.5em} \textbf{90.5} \hspace{0.5em} \textbf{76.8} & \textbf{89.9} & \textbf{76.7} \\
        \hline
    \end{tabular}}
    \label{tab:ab_block}
\end{table}

\textbf{Comparisons with fully-supervised methods.} Labeling dense framewise annotations for a video is significantly more time-consuming than labeling sparse timestamps. To further ensure relatively fair comparisons between two distinct supervision settings, we instead evaluate action segmentation performance under comparable annotation time budgets. Since the ratio of timestamp-supervised videos to fully-supervised videos that can be labeled within the same time is not definitive and highly subjective, we employ an approximate comparison by reducing the number of fully-supervised training videos by half. Specifically, for fully-supervised methods, we randomly select 50\% of the training videos for network learning while ensuring coverage of all action classes, and maintaining the rest of the experimental settings unchanged. As fully-supervised methods cannot utilize unlabeled videos, the obtained experimental results provide an approximate comparison with our method under an equivalent annotation time constraint.

Table~\ref{tab:vs_fully} presents the original results of fully-supervised methods alongside their performance when trained on only 50\% of the videos. As the number of training videos decreases to 50\%, fully-supervised methods exhibit significantly lower performance compared to our BVN under the timestamp-supervised setting, where less than 0.5\% of frames are annotated. The experimental results not only demonstrate the effectiveness of our approach against state-of-the-art methods that rely on dense framewise annotations, but also suggest that under a fixed and limited annotation budget, annotating more videos with sparse timestamp supervision may be more beneficial than annotating fewer videos with dense full supervision.

\textbf{Computational complexity.} Table~\ref{tab:exp_computational} presents the comparison of network parameters, FLOPs, and GPU memory usage between our proposed method and Li et al.~\cite{li2021temporal}, the pioneering approach for timestamp-supervised action segmentation. The FLOPs measurements are conducted on a 2-minute video, while GPU memory usage is evaluated on the Breakfast dataset. In terms of training time, our method with ASFormer backbone requires approximately 20 hours to train on the Breakfast dataset, compared to 11 hours for Li et al.~\cite{li2021temporal}. In summary, our approach achieves notable performance improvements at the expense of increased computational complexity. However, the additional cost introduced by the voting mechanism constitutes only a small fraction compared to the impact of changing the network backbone.

\begin{table}
\centering
\caption{Effect of block numbers on the GTEA dataset.}
\scalebox{1.1}{
\begin{tabular}{c|c|c|c}
        \hline
        \textbf{Blocks num} & \textbf{F1@\{10, 25, 50\}} & \textbf{Edit} & \textbf{Acc}\\
       \hline
        1 &90.3 \hspace{0.5em} 85.8 \hspace{0.5em} 73.1 &86.5 &74.6 \\
        2 &90.9 \hspace{0.5em} 89.5 \hspace{0.5em} 75.0 &89.3 &76.2 \\
        3 &\textbf{91.7} \hspace{0.5em} \textbf{90.5} \hspace{0.5em} \textbf{76.8} & \textbf{89.9} & \textbf{76.7} \\
        \hline
    \end{tabular}}
    \label{tab:ab_num}
\end{table}

\begin{table}
\centering
\caption{Effect of $start\_net$ and $end\_net$ on the GTEA dataset.}
\scalebox{1.1}{
\begin{tabular}{c|c|c|c}
        \hline
        \textbf{Voting block} & \textbf{F1@\{10, 25, 50\}} & \textbf{Edit} & \textbf{Acc}\\
       \hline
        $start\_net$ &90.1 \hspace{0.5em} 87.7 \hspace{0.5em} 75.0 &87.6 &75.5 \\
        $end\_net$ &90.0 \hspace{0.5em} 86.5 \hspace{0.5em} 74.8 &88.5 &75.9 \\
        Both &\textbf{91.7} \hspace{0.5em} \textbf{90.5} \hspace{0.5em} \textbf{76.8} & \textbf{89.9} & \textbf{76.7} \\
        2 $start\_net$ &90.6 \hspace{0.5em} 88.8 \hspace{0.5em} 75.5 &89.1 &76.1 \\
        \hline
    \end{tabular}}
    \label{tab:ab_voting}
\end{table}


\subsection{Ablation studies}

\textbf{Baseline comparisons.}
To establish a valid baseline, we remove the proposed voting module, aggregation module, and voting loss $\mathcal{L}_{vote}$ while keeping the rest of the network architecture unchanged, as shown in Table~\ref{tab:ab_baseline}. Compared to the vanilla model, our proposed method effectively propagate and aggregate key action representations, mitigating feature ambiguity in action-transiting regions and improving overall action segmentation performance.

We further examine the framewise pseudo-label generation performance comparisons on the GTEA dataset. As shown in Table~\ref{tab:ab_global_prior}, the proposed voting mechanism significantly improves the quality of generated framewise pseudo-labels compared to the baseline, demonstrating its effectiveness in refining boundary localization and mitigating action-transiting feature ambiguity.

\begin{table*}
    \caption{Effect of the voting loss $\mathcal{L}_{vote}$ on the GTEA, 50Salads, and Breakfast datasets.}
    \label{tab:ab_loss}
    \centering
    \scalebox{1.1}{
    \begin{tabular}{l|c|c|c|c|c|c|c|c|c}
        \hline
         & \multicolumn{3}{c|}{\textbf{GTEA}} & \multicolumn{3}{c|}{\textbf{50Salads}} & \multicolumn{3}{c}{\textbf{Breakfast}} \\
        \hline
        \textbf{Settings} & F1@\{10, 25, 50\} & Edit & Acc & F1@\{10, 25, 50\} & Edit & Acc & F1@\{10, 25, 50\} & Edit & Acc \\
        \hline
        Baseline & 86.6 \hspace{0.5em} 83.0 \hspace{0.5em} 72.3 & 84.3 & 72.7 & 82.1 \hspace{0.5em} 77.8 \hspace{0.5em} 67.8 & 75.3 & 78.2 & 73.0 \hspace{0.5em} 65.6 \hspace{0.5em} 48.0 & 73.9 & 63.8 \\
        Without $\mathcal{L}_{vote}$ & 84.0 \hspace{0.5em} 78.4 \hspace{0.5em} 65.3 & 77.3 & 72.8 & 77.1 \hspace{0.5em} 74.0 \hspace{0.5em} 62.7 & 65.6 & 77.7 & 70.6 \hspace{0.5em} 61.4 \hspace{0.5em} 44.0 & 69.3 & 62.8 \\
        \textbf{With $\mathcal{L}_{vote}$} & \textbf{91.7} \hspace{0.5em} \textbf{90.5} \hspace{0.5em} \textbf{76.8} & \textbf{89.9} & \textbf{76.7} & \textbf{85.1} \hspace{0.5em} \textbf{82.5} \hspace{0.5em} \textbf{72.3} & \textbf{79.0} & \textbf{81.2} & \textbf{77.3} \hspace{0.5em} \textbf{69.7} \hspace{0.5em} \textbf{51.2} & \textbf{76.5} & \textbf{68.3} \\
        \hline
    \end{tabular}}
\end{table*}

\begin{table}
\centering
\caption{Effect of hyper-parameters $\alpha$, $\beta$, and $\gamma$ on the GTEA dataset.}
\scalebox{1.1}{
\begin{tabular}{c|c|c|c}
        \hline
        $\bm{\alpha}$ & \textbf{F1@\{10, 25, 50\}} & \textbf{Edit} & \textbf{Acc}\\
       \hline
        0.05 &89.2 \hspace{0.5em} 88.2 \hspace{0.5em} 74.4 &88.4 &75.0 \\
        0.1 &90.8 \hspace{0.5em} 89.5 \hspace{0.5em} 75.4 &88.7 &75.6 \\
        \textbf{0.15} &\textbf{91.7} \hspace{0.5em} \textbf{90.5} \hspace{0.5em} \textbf{76.8} & \textbf{89.9} & \textbf{76.7} \\
        0.25  &90.2 \hspace{0.5em} 89.0 \hspace{0.5em} 75.4 &89.1 &75.2 \\
        \hline
        $\bm{\beta}$ & \textbf{F1@\{10, 25, 50\}} & \textbf{Edit} & \textbf{Acc}\\
       \hline
        0.025  &89.0 \hspace{0.5em} 86.4 \hspace{0.5em} 73.6 & 87.1 & 74.0 \\
        0.05  &89.4 \hspace{0.5em} 88.2 \hspace{0.5em} 75.1 & 89.0 & 75.2 \\
        \textbf{0.075} &\textbf{91.7} \hspace{0.5em} 90.5 \hspace{0.5em} \textbf{76.8} & \textbf{89.9} & \textbf{76.7} \\
        0.1   &91.4 \hspace{0.5em} \textbf{90.7} \hspace{0.5em} 76.0 & 89.4 & 76.3 \\
        \hline
        $\bm{\gamma}$ & \textbf{F1@\{10, 25, 50\}} & \textbf{Edit} & \textbf{Acc}\\
       \hline
        0.005 &88.6 \hspace{0.5em} 86.8 \hspace{0.5em} 73.5 & 86.3 & 74.1 \\
        \textbf{0.01}  &\textbf{91.7} \hspace{0.5em} \textbf{90.5} \hspace{0.5em} 76.8 & \textbf{89.9} & \textbf{76.7} \\
        0.015 &91.5 \hspace{0.5em} 89.6 \hspace{0.5em} \textbf{77.2} & 88.6 & 76.4 \\
        0.02 &90.6 \hspace{0.5em} 89.0 \hspace{0.5em} 76.0 & 88.2 & 75.8 \\
        \hline
    \end{tabular}}
    \label{tab:ab_alpha}
\end{table}

\begin{table}
\centering
\caption{Effect of distance threshold $E$ on the GTEA dataset.}
\scalebox{1.0}{
\begin{tabular}{c|c|c|c}
        \hline
        \textbf{$E$} & \textbf{F1@\{10, 25, 50\}} & \textbf{Edit} & \textbf{Acc}\\
       \hline
        0    &89.1 \hspace{0.5em} 87.4 \hspace{0.5em} 74.3 & 87.7 & 74.7 \\
        0.05 &90.6 \hspace{0.5em} 89.7 \hspace{0.5em} 76.1 & 88.3 & 76.2 \\
        \textbf{0.1}  &\textbf{91.7} \hspace{0.5em} \textbf{90.5} \hspace{0.5em} \textbf{76.8} & \textbf{89.9} & \textbf{76.7} \\
        0.15 &91.4 \hspace{0.5em} 90.1 \hspace{0.5em} 76.2 & 89.6 & 76.5 \\
        \hline
    \end{tabular}}
    \label{tab:ab_threshold_E}
\end{table}

\textbf{Voting and aggregation blocks.}
Separately analyzing the performance impact of the proposed voting and aggregation blocks is challenging. First, retaining only the voting block while removing the aggregation block renders the generated votes ineffective, as they do not influence the original video features despite capturing key action representations. Second, isolating the aggregation block is infeasible, as it requires votes from the voting block as input.

To the best of our knowledge, we conduct ablation studies with the voting block alone and assess the impact of the aggregation block by evaluating different aggregation approaches. As shown in Table~\ref{tab:ab_block}, applying only the voting block results in action segmentation performance comparable to the baseline. Regarding different aggregation strategies, our current design, which combines MLP and max pooling, significantly outperforms direct max pooling of vote features and proves to be more effective than average pooling.

\textbf{Number of blocks.} Our method sequentially processes video features through multiple stages, including the voting block, aggregation block, and decoder. To further demonstrate the effectiveness of the proposed blocks, we maintain the decoder number at 3 while varying the number of voting and aggregation blocks to 1, 2, and 3. We incorporate these blocks before the final decoder for configuration 1 and before the second and final decoders for configuration 2. As shown in Table~\ref{tab:ab_num}, BVN achieves optimal performance when both the voting and aggregation blocks are integrated before every decoder, demonstrating the necessity of hierarchical global prior knowledge propagation.

\begin{table}
    \centering
    \caption{Effect of key vote group numbers on the GTEA dataset.}
    \scalebox{1.1}{
    \begin{tabular}{c|c|c|c}
        \hline
        \textbf{Groups num} & \textbf{F1@\{10, 25, 50\}} & \textbf{Edit} & \textbf{Acc}\\
       \hline
        4 &88.6 \hspace{0.5em} 85.3 \hspace{0.5em} 74.5 &85.8 &73.3 \\
        8 &89.7 \hspace{0.5em} 87.5 \hspace{0.5em} 75.5 &87.0 &75.9 \\
        16 &\textbf{91.7} \hspace{0.5em} \textbf{90.5} \hspace{0.5em} \textbf{76.8} & \textbf{89.9} & \textbf{76.7} \\
        32 &88.2 \hspace{0.5em} 85.6 \hspace{0.5em} 74.7 &85.3 &74.2 \\
        \hline
    \end{tabular}}
    \label{tab:ab_group}
\end{table}

\begin{table}
\centering
\caption{Effect of varying action length on the GTEA dataset.}
\scalebox{1.1}{
\begin{tabular}{c|c|c|c}
        \hline
        \textbf{Sampling Scales} & \textbf{F1@\{10, 25, 50\}} & \textbf{Edit} & \textbf{Acc}\\
       \hline
        $\times$0.25 &\textbf{91.8} \hspace{0.5em} \textbf{90.6} \hspace{0.5em} 76.7 &89.3 &76.5 \\
        $\times$0.5  &91.6 \hspace{0.5em} 90.4 \hspace{0.5em} 76.5 &89.6 &76.4 \\
        $\times$1 (Original)  &91.7 \hspace{0.5em} 90.5 \hspace{0.5em} \textbf{76.8} & \textbf{89.9} & \textbf{76.7} \\
        $\times$2    &91.4 \hspace{0.5em} 90.3 \hspace{0.5em} 76.3 &89.5 &76.4 \\
        \hline
    \end{tabular}}
    \label{tab:ab_vary}
\end{table}

\textbf{$\bm{start\_net}$ and $\bm{end\_net}$.} Within each voting block, we employ $start\_net$ and $end\_net$ to direct votes towards the action-transiting regions before and after the action. To evaluate the effectiveness of these two independent networks, Table~\ref{tab:ab_voting} presents the action segmentation results using $start\_net$ only, $end\_net$ only, and both two deep networks in the voting block. Despite the notable performance achieved by each network, the combination of both $start\_net$ and $end\_net$ further significantly enhances the overall performance. Drawing upon the current experimental findings, no conclusive evidence indicates whether $start\_net$ or $end\_net$ exhibits superior performance when presented individually.

It is worth noting that incorporating both $start\_net$ and $end\_net$ inevitably increases the model parameters compared to applying a single module. To mitigate this effect, we duplicate two identical $start\_net$ modules, with the results shown in Table~\ref{tab:ab_voting}. While the inclusion of two $start\_net$ modules yields a slight performance improvement, BVN achieves state-of-the-art results when both $start\_net$ and $end\_net$ are integrated. In summary, by voting ahead and behind into the action-transiting regions, frames adjacent to the boundary obtain global prior knowledge from different perspectives, thereby collaboratively contributing to effective video frame representation learning.

\textbf{Voting loss.} The voting loss, $\mathcal{L}_{vote}$, plays a crucial role in the network's learning process. To evaluate its effectiveness, we report results for the ``Without $\mathcal{L}_{vote}$" setting in Table~\ref{tab:ab_loss}, where $\mathcal{L}_{vote}$ is excluded from the final loss function. We employ the same baseline setting as in Table~\ref{tab:ab_baseline}.

When the voting loss is omitted during training, performance declines significantly across all datasets as expected. The deterioration occurs because the absence of $\mathcal{L}_{vote}$ disrupts the voting block's ability to propagate global prior knowledge effectively into action-transiting regions. Without guidance, the voting process becomes random, invalidating the subsequent aggregation step. As a result, the absence of the voting loss introduces noise into the original video features, impairing the learning process compared to the baseline results.

\textbf{$\bm{\alpha}$, $\bm{\beta}$, and $\bm{\gamma}$.} The hyperparameters $\alpha$, $\beta$, and $\gamma$ control the scaling of $\mathcal{L}_{smooth}$, $\mathcal{L}_{conf}$, and $\mathcal{L}_{vote}$ in the loss function. As shown in Table~\ref{tab:ab_alpha}, the optimal performance is achieved with $\alpha = 0.15$, $\beta = 0.075$, and $\gamma = 0.01$, with the numerical selection of hyperparameters primarily based on empirical observations. For the experimental results on each hyperparameter, the remaining hyperparameters are set to their default optimal values. 


\textbf{Distance threshold.} The distance threshold $E$ impacts the overall model performance. As shown in Table~\ref{tab:ab_threshold_E}, an excessively low threshold may fail to propagate sufficient global prior knowledge from representative votes to action-transiting regions, whereas a high threshold may introduce irrelevant votes and noise, both of which lead to performance degradation. A threshold of 0 indicates that only knowledge from key votes is aggregated into the original video features.

\textbf{Number of key vote groups.} The number of key vote groups within the aggregation block, denoted as $P$, impacts the overall model performance. When aggregating all votes, an insufficient value of $P$ fails to cover action-transiting regions throughout the video. In contrast, an excessively large value of $P$ may tend to introduce noises during the global prior knowledge propagation procedure. Consequently, we investigate the influence of the group number $P$ on the model performance in Table~\ref{tab:ab_group}. The increase of $P$ in the early stages results in a continuous enhancement of our model performance on the GTEA dataset. However, raising $P$ from 16 to 32 leads to performance deterioration, attributed to the inclusion of excessive noises within key vote groups beyond the action-transiting regions.

\textbf{Varying action lengths.} To evaluate the generalization capability for actions of varying lengths, we modify the testing video dataset by downsampling and upsampling the video features at different scales. For instance, the $\times$0.5 downsampling in Table~\ref{tab:ab_vary} refers to selecting one frame out of every two consecutive frames, effectively halving the action length while preserving the content. Conversely, the $\times$2 upsampling interpolates an additional frame between every two consecutive frames, doubling the action length. By testing the trained model on videos with varying sampling scales, we assess its generalization capability to actions of different lengths while ensuring a fair comparison. As shown in Table~\ref{tab:ab_vary}, our proposed method maintains reliable performance across action lengths ranging from 0.25 to 2 times the original scale. In summary, our method demonstrates strong generalization capability for actions of varying lengths.

\section{Conclusions}
In this paper, we have introduced a global-to-local boundary voting network, named BVN, for timestamp-supervised action segmentation. The BVN proposes a voting mechanism by rearranging the globally evenly distributed temporal feature series and aggregating key action representations into local action-transiting regions. Through hierarchical propagation, BVN continuously addresses inherent action-transiting feature ambiguity and suppresses the ambiguous regions, leading to refined boundary localization, which proves to be essential for segmentation model training. With extensive experimental results on three public datasets, BVN has achieved state-of-the-art performance on the action segmentation task across multiple evaluation metrics.

\section{Acknowledgement}
This work was supported in part by the National Science Foundation of China under Grant 62206147, and by the National Research Foundation, Singapore, under the NRF Medium-Sized Centre Scheme (CARTIN). Any opinions, findings, and conclusions in this material are those of the authors and do not reflect the views of the National Science Foundation of China or the National Research Foundation, Singapore.

\bibliographystyle{IEEEtran}
\bibliography{egbib}

\newpage


\begin{IEEEbiography}[{\includegraphics[width=1in,height=1.25in,clip,keepaspectratio]{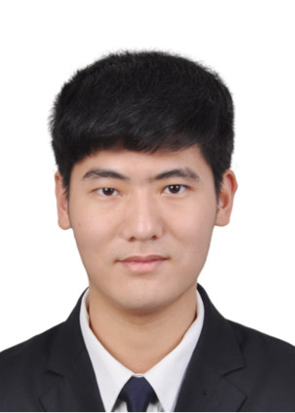}}] {Runzhong Zhang} is a PhD candidate at the School of Electrical and Electronic Engineering, Nanyang Technological University, Singapore. He obtained his B.S. degree from Xi'an Jiaotong University in 2018 and his M.S. degree from Columbia University in 2020. His current research interests include video understanding, human action analysis, and label-efficient learning.
\end{IEEEbiography}

\begin{IEEEbiography}[{\includegraphics[width=1in,height=1.25in,clip,keepaspectratio]{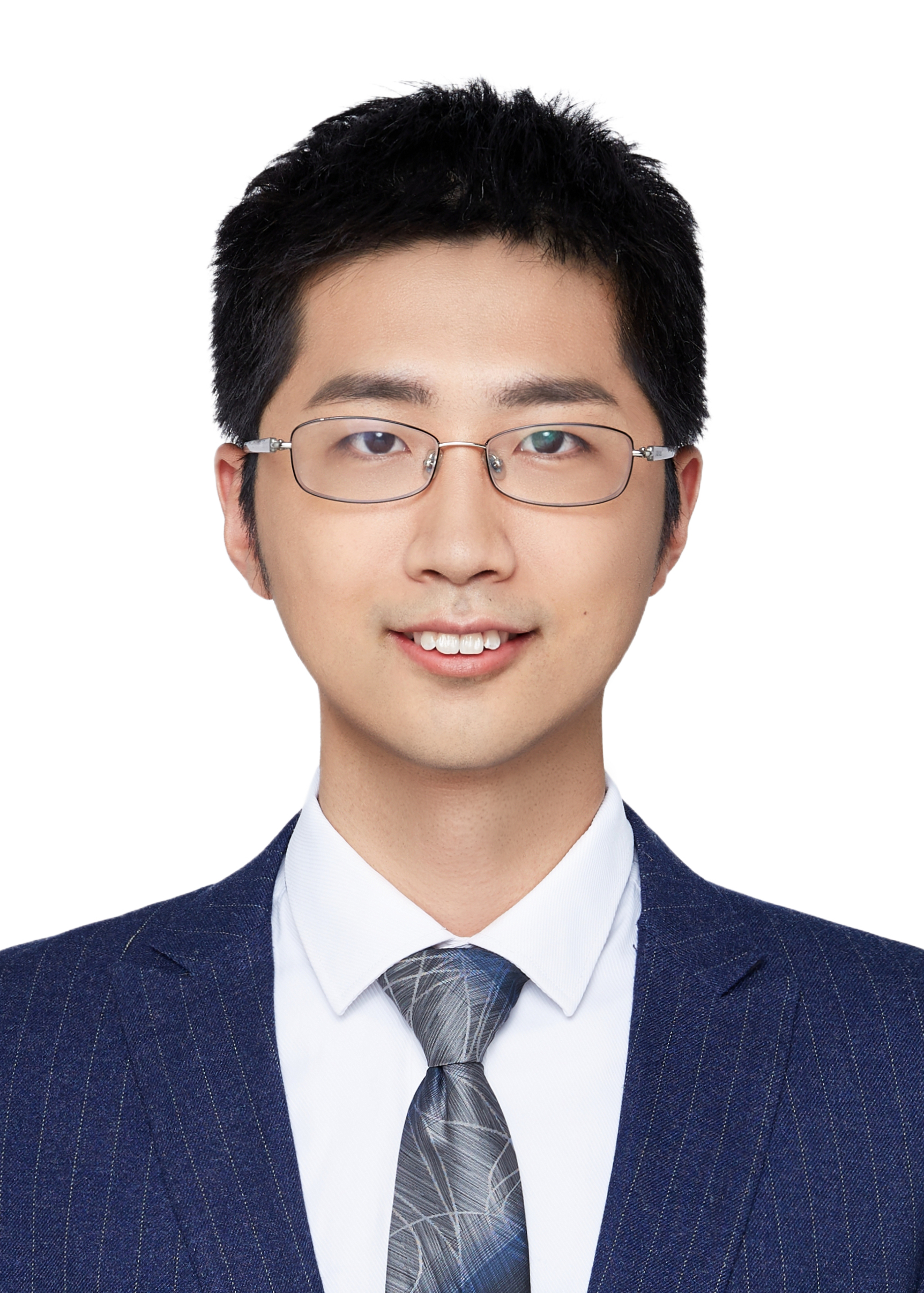}}]{Yueqi Duan} (Member, IEEE) received the B.S. and Ph.D. degrees from the Department of Automation, Tsinghua University, in 2014 and 2019, respectively. He is currently an Assistant Professor with the Department of Electronic Engineering, Tsinghua University. He has published more than 30 scientific papers in the top journals and conferences, including IEEE TRANSACTIONS ON PATTERN ANALYSIS AND MACHINE INTELLIGENCE, IEEE TRANSACTIONS ON IMAGE PROCESSING, CVPR, ICCV, ECCV and NeurIPS. His research interests include computer vision and pattern recognition. He served as the Publication Chair for FG, the Area Chair for CVPR, ICLR, MM and ICME, and a Regular Reviewer for a number of journals and conferences, e.g., IEEE TRANSACTIONS ON PATTERN ANALYSIS AND MACHINE INTELLIGENCE, IEEE TRANSACTIONS ON IMAGE PROCESSING, IJCV, CVPR, ICCV, ECCV, ICML, NeurIPS, and SIGGRAPH. He was awarded the Excellent Doctoral Dissertation of Chinese Association for Artificial Intelligence (CAAI) in 2020.
\end{IEEEbiography}

\begin{IEEEbiography}[{\includegraphics[width=1in,height=1.25in,clip,keepaspectratio]{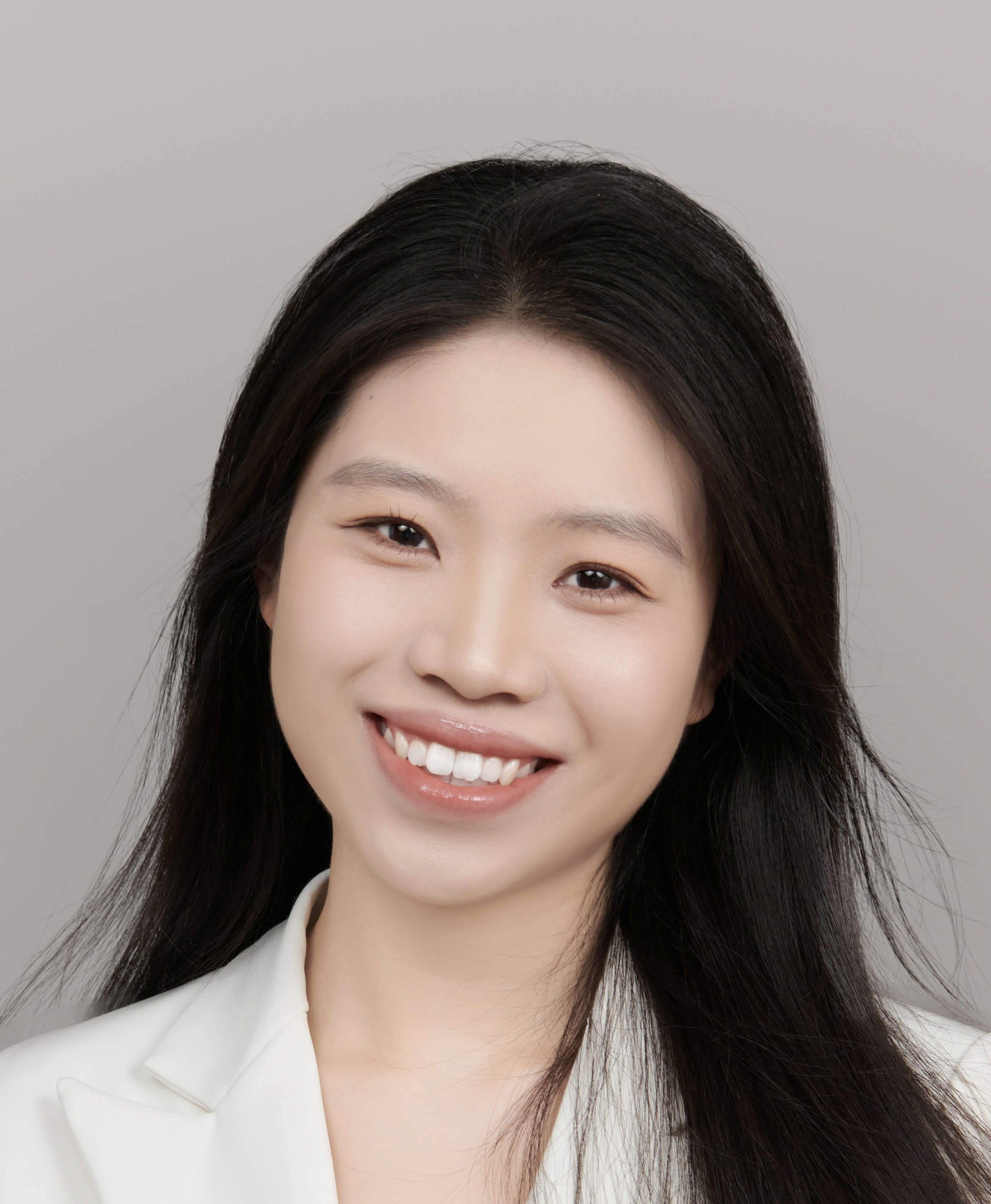}}] {Yang Chen} (Graduate Student Member, IEEE) received the Master's degree from the University of Michigan, Ann Arbor, the United States, in 2022. She is currently a Ph.D. candidate at the School of Electrical and Electronic Engineering, Nanyang Technological University, Singapore. Her current research interest is 3D vision.
\end{IEEEbiography}

\begin{IEEEbiography}[{\includegraphics[width=1in,height=1.25in,clip,keepaspectratio]{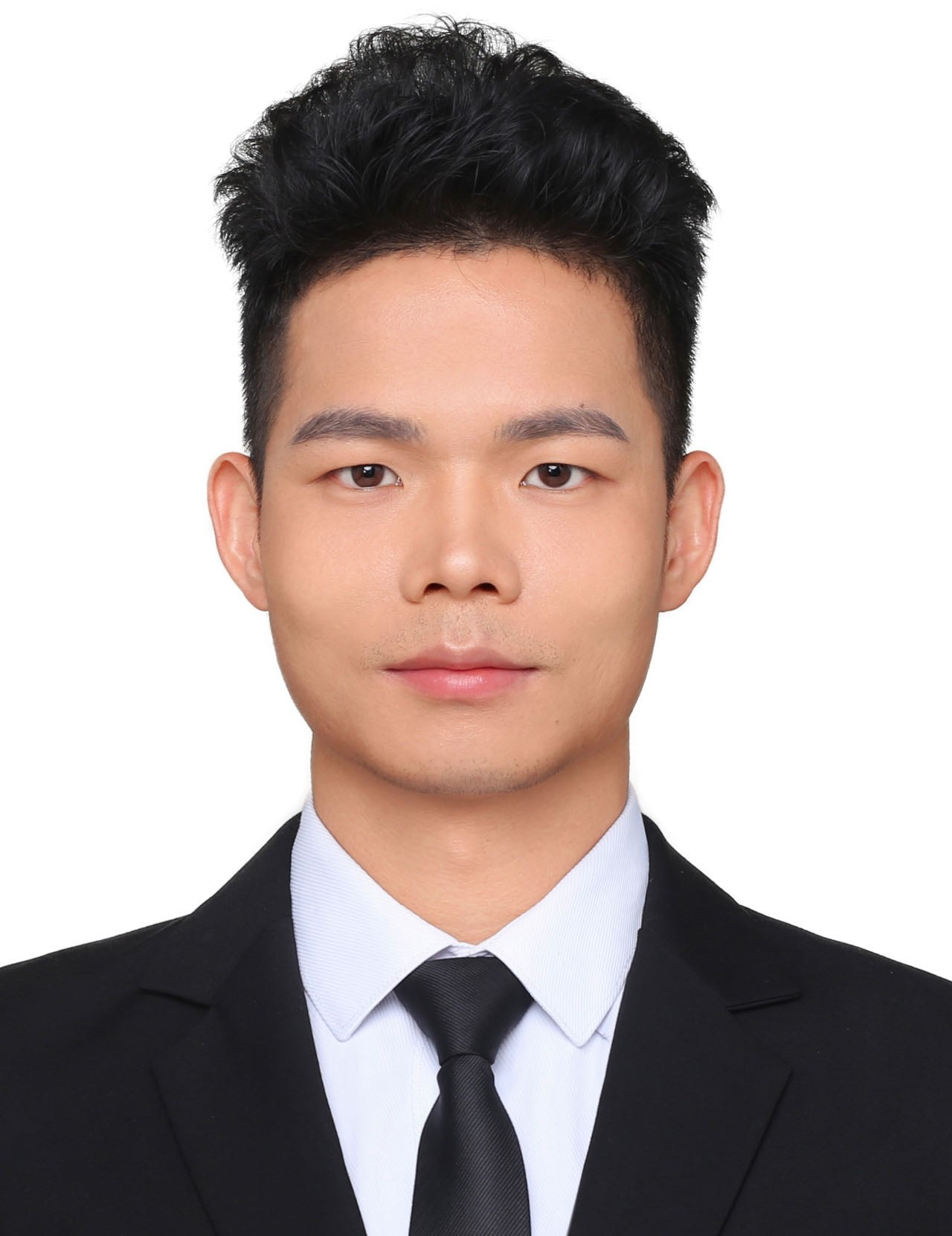}}] {Weipeng Hu}
received the Ph.D. degree in Electronics and Information Technology from Sun Yat-sen University, Guangzhou, China, in 2022. He is currently a Research Fellow with the Centre for Advanced Robotics Technology Innovation Laboratory, School of Electrical and Electronic Engineering, Nanyang Technological University, Singapore. His current research interests include image \& video synthesis, human-robot interaction, heterogeneous face recognition, and cross-domain person ReID.
\end{IEEEbiography}

\begin{IEEEbiography}[{\includegraphics[width=1in,height=1.25in,clip,keepaspectratio]{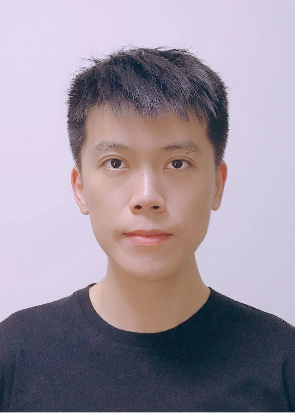}}] {Chen Cai} received his B.S. degree in Electrical and Computer Engineering from the National University of Singapore. He is currently a Ph.D. candidate at the School of Electrical and Electronic Engineering, Nanyang Technological University, Singapore. His research interests include computer vision, multimodal learning, and machine learning.
\end{IEEEbiography}

\begin{IEEEbiography}[{\includegraphics[width=1in,height=1.25in,clip,keepaspectratio]{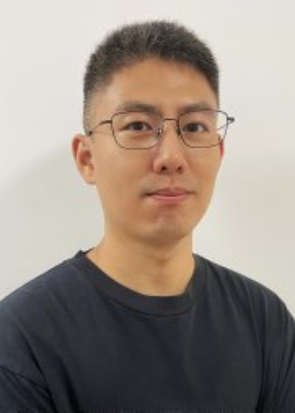}}] {Suchen Wang} received his Ph.D. degree from the School of Electrical and Electronic Engineering, Nanyang Technological University, Singapore, in 2022. He is an applied scientist at Amazon, Seattle, US. His current research interests include human-object interaction and video understanding.
\end{IEEEbiography}

\begin{IEEEbiography}[{\includegraphics[width=1in,height=1.25in,clip,keepaspectratio]{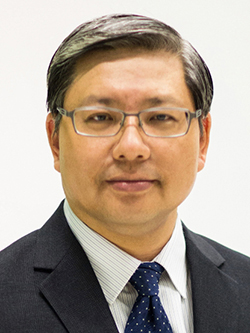}}] {Yap-Peng Tan}  (Fellow, IEEE) received the B.S. degree from National Taiwan University, Taipei, Taiwan, in 1993, and the M.A. and Ph.D. degrees from Princeton University, Princeton, NJ, in 1995 and 1997, respectively, all in electrical engineering. He is currently a Professor and Associate Vice President at Nanyang Technological University (NTU), Singapore. His
research interests include image and video processing, machine learning, computer vision, and data analytics. He served as an Associate Editor of the IEEE TRANSACTIONS ON CIRCUITS AND SYSTEMS FOR VIDEO TECHNOLOGY, IEEE SIGNAL PROCESSING LETTERS, IEEE TRANSACTIONS ON MULTIMEDIA, and IEEE Access, as well as an Editorial Board Member of the EURASIP Journal on Advances in Signal Processing and EURASIP Journal on Image and Video Processing. He was the Technical Program Co-Chair of the 2015 IEEE International Conference on Multimedia and Expo (ICME 2015) and the 2019 IEEE International Conference on Image Processing (ICIP 2019), and the General Co-Chair of the 2010 IEEE International Conference on Multimedia and Expo (ICME 2010) and the 2015 IEEE International Conference on Visual Communications and Image Processing (VCIP 2015).
\end{IEEEbiography}

\end{document}